\pdfoutput=1

\documentclass[10pt,journal,compsoc]{IEEEtran}

\usepackage{times}
\usepackage{epsfig}
\usepackage{graphicx}
\usepackage{amsmath}
\usepackage{amssymb}
\usepackage{multirow}
\usepackage{enumitem}
\usepackage{xcolor,colortbl}
\usepackage{booktabs}
\usepackage{afterpage}
\usepackage[lofdepth,lotdepth]{subfig}
\usepackage[bookmarks=false]{hyperref}
\usepackage{tikz}
\usetikzlibrary{arrows}
\usetikzlibrary{shapes}
\usepackage[normalem]{ulem}
\usepackage{review}

\def\eg{\textit{e.g. }}
\def\ie{\textit{i.e.}}

\def\etal{\textit{et al.}}

\newcommand{\printfnsymbol}[1]{%
  \textsuperscript{\@fnsymbol{#1}}%
}
\newcommand{\vect}[1]{\mathbf{#1}}

\setrevision{0}
\newcommand{\comRev}[1]{#1}
\definecolor{darkgreen}{gray}{0.0}

\ifCLASSOPTIONcompsoc
  \usepackage[nocompress]{cite}
\else
  \usepackage{cite}
\fi
\begin{document}
%
\title{Progressive Fusion for Unsupervised Binocular Depth Estimation using Cycled Networks}
%
%
%
%

\author{Andrea~Pilzer,
        ~St\'{e}phane~Lathuili\`{e}re,
        ~Dan~Xu,
        Mihai~Marian~Puscas, ~Elisa~Ricci,~\IEEEmembership{Member,~IEEE},
        and~Nicu~Sebe,~\IEEEmembership{Senior Member,~IEEE}
\IEEEcompsocitemizethanks{\IEEEcompsocthanksitem Authors are with the Department of Information Engineering
and Computer Science, University of Trento, Trento, Italy. E-mail:
\{andrea.pilzer, stephane.lathuiliere, e.ricci, mihaimarian.puscas, niculae.sebe\}@unitn.it\protect
\IEEEcompsocthanksitem Dan Xu is with the Department of Engineering Science, University of Oxford, Oxford, UK. Email: danxu@robots.ox.ac.uk
\IEEEcompsocthanksitem Elisa Ricci is also with Fondazione Bruno Kessler. Email: eliricci@fbk.eu
\IEEEcompsocthanksitem Nicu Sebe is also with Huawei Technologies Ireland. Email: niculae.sebe@huawei.com}
\thanks{Manuscript received October 11, 2018; revised July 25, 2019.}
}

%
%

\markboth{Journal of \LaTeX}%
{Shell \MakeLowercase{\textit{et al.}}: Bare Demo of IEEEtran.cls for Computer Society Journals}
%



\IEEEtitleabstractindextext{%
\begin{abstract}
	Recent deep monocular depth estimation approaches based on supervised regression have achieved remarkable performance. However, they require costly ground truth annotations during training. To cope with this issue, in this paper we present a novel unsupervised deep learning approach for predicting depth maps. We introduce a new network architecture, named Progressive Fusion Network (PFN), that is specifically designed for binocular stereo depth estimation. This network is based on a multi-scale refinement strategy that combines the information provided by both stereo views. In addition, we propose to stack twice this network in order to form a cycle. This cycle approach can be interpreted as a form of data-augmentation since, at training time, the network learns both from the training set images (in the forward half-cycle)  but also from the synthesized images (in the backward half-cycle). The architecture is
	jointly trained with adversarial learning. Extensive experiments on the publicly available datasets KITTI, Cityscapes and ApolloScape demonstrate the effectiveness of the proposed model which is competitive with other unsupervised deep learning methods for depth prediction.

\end{abstract}

}

\maketitle
\IEEEdisplaynontitleabstractindextext

%
\IEEEpeerreviewmaketitle


%
%
%
%
\renewcommand{\arraystretch}{1.1}
\section{Introduction}
\label{sec:intro}



Most previous works considering deep architectures for predicting depth maps operate in a supervised learning setting~\cite{eigen2014depth,ladicky2014pulling,liu2016learning,xu2018monocular} and employ powerful deep regression models based on Convolutional Neural Networks (ConvNet). These models are usually designed for monocular depth estimation, \textit{i.e.} they 
are trained to learn the transformation from a single RGB image to a depth map in a pixel-to-pixel fashion.  
However, supervised learning models require ground-truth depth data which are usually costly to acquire. This problem is especially relevant
with deep learning architectures, as large amounts of data are typically required in order to produce satisfactory performance. 
Furthermore, monocular depth estimation is an inherently ill-posed problem due to the well-known
scale ambiguity issue~\cite{saxena2006learning}.  For instance, 
given an image patch of a blue sky, it is difficult to predict if this patch is infinitely far away (sky), or whether it is part of a blue object. Therefore, local information such as the texture must be combined with contextual information. Additionally, in complex environments as those encountered by autonomous driving cars, current depth estimation methods still have difficulties in predicting accurately depth maps from a single camera. These difficulties are encountered in particular when many objects are present in the scene due to the several occlusions. 

\begin{figure}[!t]
	\includegraphics[width=0.5\columnwidth,angle=-90]{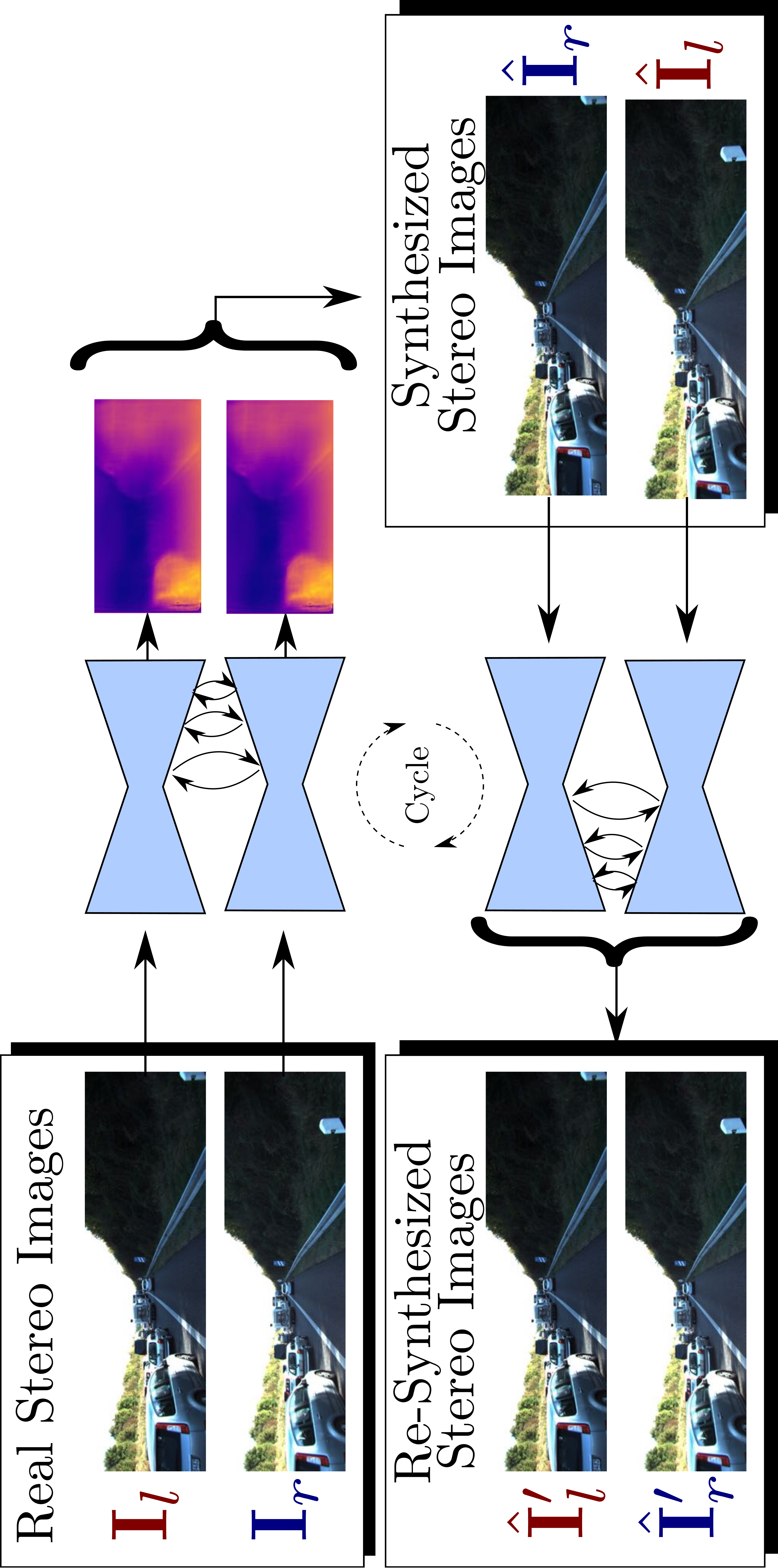}
	\caption{Motivation of the proposed unsupervised depth
          estimation approach using cycled generative networks. 
          }
	\label{fig:motivation}
	\vspace{-15pt}
\end{figure}


To tackle these problems, \textit{unsupervised} (also called self-supervised) learning-based approaches for depth estimation operating on a \textit{stereo setting} have been introduced~\cite{luo2016efficient,mayer2016large,godard2017unsupervised,wang2017learning}.
These methods operate by learning the correspondence field (\ie~the disparity map) between two different image views of a calibrated stereo camera using only the left and right RGB images (no ground-truth depth map).
The disparity refers to the difference in image location of an object seen by the left and right cameras. Importantly, the disparity value is inversely proportional to the object depth at the corresponding pixel location. Then, given the calibration parameters of the stereo cameras, the depth maps can be calculated using the predicted disparity maps.
At test time, depending on the network architecture, depth is estimated either from a single stereo view \cite{godard2017unsupervised} (referred to as \emph{monocular depth estimation}) or 
 stereo pairs\cite{pilzer2018unsupervised} (referred to as \emph{binocular} or \emph{stereo depth estimation}). Thanks to this formulation, we avoid ground-truth data collection, using lidar for instance, that is much more complex (eg. multimodal-synchronization, hardware constraints) and expensive than adding a second camera. Another potential advantage of unsupervised depth estimation can be found in online adaptation as in \cite{tonioni2019cvpr}, where the network is self-adapted in an online fashion at testing time when depth supervision is no more available.~
Most of previous works~\cite{godard2017unsupervised,wang2017learning} are based on a common strategy introduced in \cite{garg2016unsupervised}: given a pair of left and right images a neural network is trained for the task of predicting
the right-to-left disparity map from the left image. The left image can then be re-synthesized by warping the right image accordingly to the predicted disparity. The network is trained by minimizing a left image reconstruction loss (see Sec~\ref{sec:pbstatement} for technical details). This approach relies on the 
supervision from the image synthesis of an expected view, whose quality plays a direct influence on the performance of the estimated disparity map.  
Significant progress has been made recently along this research line \cite{godard2017unsupervised,zhou2017unsupervised,mahjourian2018unsupervised}.

In this paper, we follow this research thread 
and propose a novel end-to-end trainable deep network model for adversarial learning-based depth estimation given stereo image pairs. Contrary to most recent works~\cite{godard2017unsupervised,wang2017learning,garg2016unsupervised}, we focus on the binocular scenario where stereo image pairs are available both at training and test time. The proposed approach consists of a generative sub-network which
predicts the two disparity maps from the right to the left views and vice-versa. 
This sub-network
is stacked twice in order to form a cycle as illustrated in Fig.~\ref{fig:motivation}. This novel 
network design provides strong constraints and supervision for each
image view, facilitating the optimization of the network.
It is important to mention that, despite the cycle shape and the use of adversarial loss, our cycle approach is not directly related to \comRev{Cycle-GAN} \cite{CycleGAN2017}. Cycle-GAN is designed for image-to-image translation when paired data are not available. In the case of binocular stereo depth estimation, paired data are available (\textit{i.e.} corresponding left/right images).
Our cycle approach can be interpreted as a form of data-augmentation
since, at training time, the network learns to predict disparity maps
from images of the training set (in the forward cycle pass), but also from
synthesized images (in the backward cycle pass). In addition, it prevents the sub-network to predict blurred or deformed images in the forward cycle pass, since 
it would suffer the consequences in the backward cycle pass.
The whole cycle is jointly learned and the final disparity map is 
produced by the first $G$ network.


Part of the material presented in this paper appeared in \cite{pilzer2018unsupervised}. The current
paper extends \cite{pilzer2018unsupervised} in several ways:
\begin{itemize}
    \item First, we present a more detailed
analysis of related works by including recently published works dealing with supervised and unsupervised depth estimation.
\item Second, we propose a novel network architecture named Progressive
Fusion Network (PFN), that is specifically designed for binocular stereo depth estimation.
This network is based on a multi-scale refinement strategy that
combines the information provided by the left and right images. 
\item Third, the cycle model proposed in \cite{pilzer2018unsupervised} is adapted in order
to benefit from the two disparity maps predicted by the proposed PFN.
\item Finally, we significantly extend our quantitative evaluation by performing an in-detail 
ablation study and by comparing our binocular
stereo model with the very recent works in this area. Our extensive experiments on three large publicly available datasets (\ie~KITTI \cite{Geiger2013IJRR}, Cityscapes \cite{Cordts2016Cityscapes} and ApolloScape\cite{apollo}) demonstrate the effectiveness of the proposed
adversarial image synthesis, cycled generative network structure
and Progressive Fusion Network. On the widely used KITTI dataset, our approach is competitive with state of the art methods on the static unsupervised setting.
\end{itemize}

The rest of the paper is organized as follows. In  Section~\ref{sec:related}, we
analyze the related work considering both supervised and unsupervised depth
estimation methods. The details of our method are presented in Section~\ref{sec:method}.
Section \ref{sec:exp} presents the experimental evaluation and conclusions
are drawn in Section \ref{sec:concl}.

\section{Related Work}
\label{sec:related}
In this section we review previous works focusing on depth estimation with deep learning models and, more broadly, considering pixel-level prediction tasks. We also briefly discuss previous approaches on cascade regression as the proposed PFN strategy is inspired from them.

\subsection{Supervised Depth Estimation} In the last decade, deep learning models have greatly
improved the performance of supervised approaches for depth estimation.  Given enough training data, 
deep neural networks have been shown to be especially effective in predicting depth maps in a monocular setting 
~\cite{eigen2014depth,liu2016learning,zhuo2015indoor,laina2016deeper,xu2018monocular}. A
first ConvNet was proposed by Eigen \etal~\cite{eigen2014depth}, where the benefit of considering both local and global information was demonstrated. Other works considered
probabilistic graphical models implemented
as neural networks for end-to-end optimization, boosting the performance of deep regression models
\cite{liu2016learning,wang2015towards,xu2018structured,xu2018monocular}. 
In particular, Wang \etal \cite{wang2015towards} proposed a ConvNet integrated with a hierarchical Conditional Random Fields (CRFs) for joint depth estimation and semantic segmentation. 
Xu~\etal~\cite{xu2018structured} introduced a CRFs-based approach for learning deep representations and highlighted the benefit of exploiting information from multiple scales to improve depth predictions. A similar idea was also exploited in \cite{xu2018monocular}, where a structured attention mechanism was integrated into the CRFs. 
Similarly, stereo matching has been tackled using supervised ConvNet models. Chang et. al~\cite{chang2018pyramid} introduce a pyramid pooling module for incorporating global context information into image features. In~\cite{pang2017cascade} a residual refinement network is adopted to to improve prediction. Finally, Tulyakov et al.~\cite{tulyakov2018practical} propose a sub-pixel cross-entropy loss combined with a MAP estimator in order to handle stereo ambiguous matches.
However, supervised learning-based approaches rely on expensive ground-truth depth data for training and are not flexible to be deployed in novel environments. Even if synthetic data generation has been proposed to partially tackle this issue \cite{atapour2018real}, the cost of synthesizing realistic data remains high.


\subsection{Unsupervised Depth Estimation} More recent works proposed deep learning models for unsupervised learning-based depth estimation~\cite{kuznietsov2017semi,mahjourian2018unsupervised,wang2017learning,zhan2018unsupervised,pilzer2019cvpr}, thus avoiding the use of costly ground truth depth annotations. For instance, Garg~\etal~\cite{garg2016unsupervised} introduced an approach to learn predicting the depth map in an indirect way. They used a Convolutional Neural Network to estimate the right-to-left disparity map from the left image and then to accordingly warp the right image to the predicted disparity in order to reconstruct the left image. They also trained the network in order to minimize the discrepancy between the original and the reconstructed left image. Improving upon \cite{garg2016unsupervised}, Godard~\etal~\cite{godard2017unsupervised} 
proposed to estimate both the left-to-right and right-to-left disparity maps using a single generative network and utilized the consistency between them to constrain the model. In~\cite{zhong2017self}, 3D convolutions are employed to perform 3D feature matching leading to computationally costly model (inference about 1 fps).~  Zhou~\etal~\cite{zhou2017unsupervised} introduced an approach to jointly learn the depth and the camera pose adopting a single deep network.
Importantly, most of these works focus on the monocular setting since, at test time, depth is estimated from a single stereo view. Conversely, in this work, we focus on the binocular stereo depth estimation scenario where both images are available at test time.
Other works considered jointly learning the scene depth and the ego-motion in monocular videos without using ground-truth data~\cite{wang2018learning,yang2018every,Yang2018ECCV,godard2018digging,zhou2017unsupervised}. 
These works demonstrated that by integrating temporal information and considering multiple consecutive frames better estimates can be obtained.
In opposition to these works, we purely focus on improving the performance of unsupervised frame-level depth estimation without exploiting any additional information such as temporal consistency or supervision from related tasks (\eg     segmentation, ego-motion estimation, etc.). 


\subsection{Pixel-level prediction}
Beyond depth estimation, great progress has  also been made in many pixel-level prediction tasks thanks to the development of deep learning architectures and, in particular, of fully convolutional neural networks. Specifically, pixel-level prediction tasks refer to estimating a continuous value or a category label for each pixel of an input image. Notable examples are semantic segmentation\cite{luc2016semantic,xue2018segan}, surface normal estimation~\cite{wang2018learning},
denoising \cite{xie2012image} or image colorization\cite{zhang2016colorful}. Some of these works also stressed the importance of considering multi-scale deep representations for improving accuracy \cite{chen2016attention,porzi2017learning}.
Importantly, many pixel-level prediction methods employed a U-Net architecture \cite{ronneberger2015u}. This architecture consists of an encoder-decoder network that uses skip connections to preserve local information. However, this architecture is designed for the task in which the input and the output images are geometrically aligned \cite{siarohin2018deformable}. In this paper we show that, since binocular stereo depth estimation suffers from the misalignment problem (see Sec.~\ref{sec:CFN}), the U-Net architecture is insufficient and requires some adaptation. 

Recently, Generative Adversarial Networks (GANs) have attracted a lot of attention, not only 
for image generation tasks~\cite{chen2017photographic} but also for pixel-level prediction problems~\cite{isola2017image}. A major advantage of GANs is that they consider a global consistency loss oppositely to traditional losses such as $\mathcal{L}_1$ or $\mathcal{L}_2$ losses that act only at pixel level. For instance, in the case of image segmentation, Luc \etal~\cite{luc2016semantic} showed that adversarial training better enforces spatial consistency among the class labels. In this paper we argue that spatial consistency is also needed in depth estimation and propose an adversarial learning scheme for training depth prediction networks.

Among all pixel-level prediction tasks,
optical flow estimation is surely the most closely related to binocular stereo depth estimation. 
On one hand, optical flow estimates the 2D shift of each pixel to align one image at time $t$ and the following image at time $t+1$ or vice-versa. On the other hand, unsupervised binocular stereo depth estimation consists in learning to estimate a disparity map (horizontal pixel shift) to align the two stereo views at the same time $t$. In~\cite{flownet}, a convNet named \emph{flowNet} is supervisingly trained to predict optical flow~\cite{flownet}. More recently, a self-supervised approach was proposed to implement self-supervised optical flow estimation~\cite{Meister2018unflow}. Although the employed self-supervised formulation is close to our approach, we employ a specific coarse to fine estimation strategy.

\subsection{Coarse to fine estimation}

  Many regression problems in computer vision have been addressed with a \emph{coarse to fine} estimation approach. For regression problems, the coarse to fine strategy is often implemented via cascade regression~\cite{chen2017photographic,dollar2010cascaded}: a first coarse estimation of the values of interest is combined with the input image in order to refine the prediction. The refinement procedure is repeated until convergence.~This approach provides good results since it combines the global information encoded in the current estimation with the local information present in the image.
A standard implementation, as in \emph{flowNet}~\cite{flownet} consists in stacking several ConvNets resulting in a high computation cost.
\addnote[continuous-fashion]{1}{To keep a low computation cost, we combine the coarse-to-fine strategy  with a multi-scale learning framework that helps to iteratively fuse the information from multiple cameras from low to high resolution}. In our proposed model, an estimation is first performed at low resolution, and it is then used to tackle the misalignment problem and further refine the prediction. Therefore, the estimation is gradually refined within a single decoder network.


\section{Proposed Approach}
\label{sec:method}

As mentioned in the introduction, our framework for unsupervised depth
estimation has two main contributions. First, we propose to exploit cycle consistency in order to regularize better our model and therefore achieve better performance. The cycle consistency approach is combined with an adversarial learning strategy in order to further improve the predictions. The details of our model are given in Sections \ref{sec:USDE} and \ref{sec:cycle}. Second, we propose a network architecture named Progressive Fusion Network, that uses a multi-scale approach to
fuse the information provided by each image. The motivations and the details of this network architecture are given in Section~\ref{sec:CFN}.
Before presenting the details of our approach, we briefly introduce in Sec.~\ref{sec:pbstatement} the basics of unsupervised depth estimation and the notations used in the remaining of the paper.


\subsection{Unsupervised Binocular Depth Estimation}
\label{sec:pbstatement}
\begin{figure*}[!t]
\begin{center}
\includegraphics[width=0.83\textwidth]{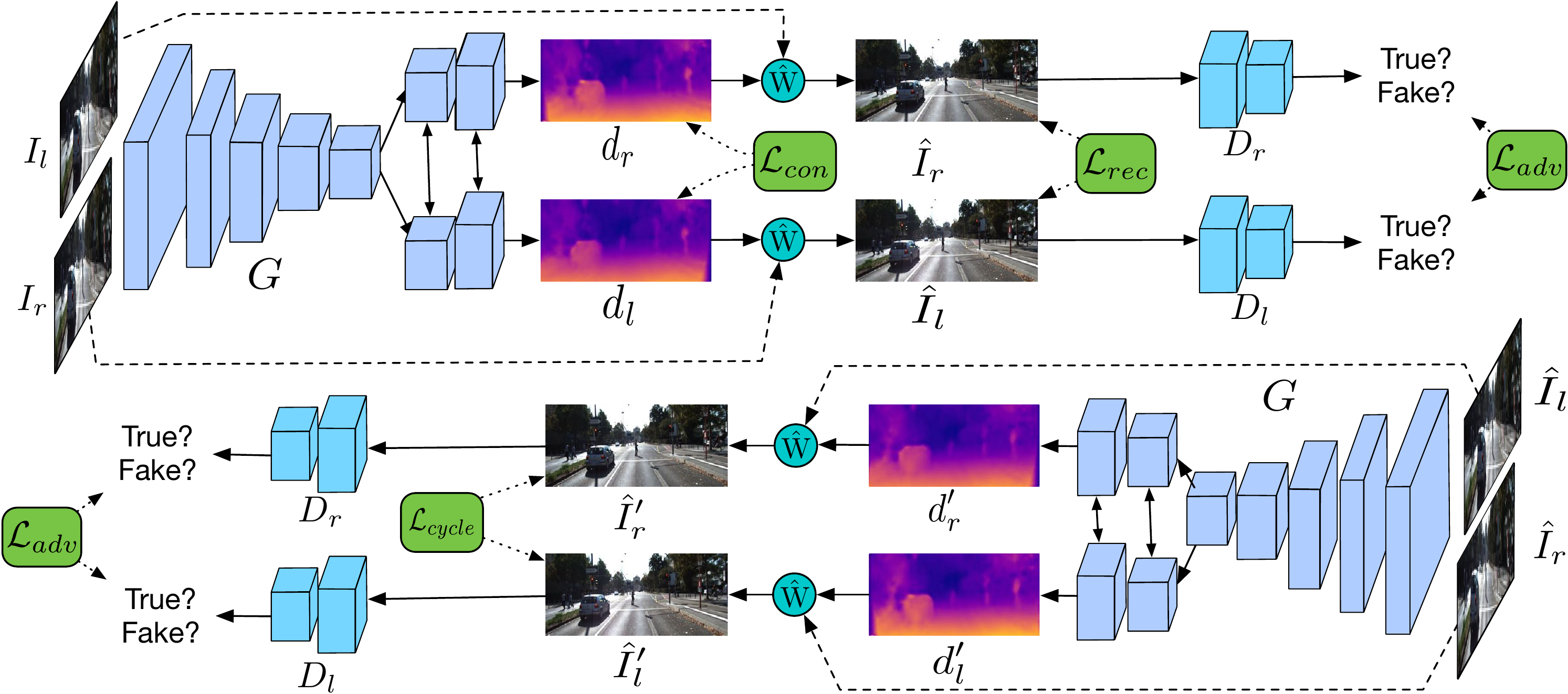}
\vspace{-4pt}
\caption{Illustration of the detailed framework of the proposed cycled generative networks with Progressive Fusion Network for unsupervised adversarial depth estimation. $\mathcal{L}_{rec}$ represents the reconstruction loss for different generators; $\mathcal{L}_{con}$ denotes a consistence loss between the disparity maps generated from the two generators.}
\label{fig:detailedgan}
\end{center}
\vspace{-10pt}
\end{figure*}

In this work, we aim at estimating a depth map given a pair of images from calibrated stereo cameras. A supervised approach would consist in learning a stereo matching network that predicts depth~\cite{luo2016efficient}. In this scenario, the network is trained via minimization of a pixel-wise error measure between the predicted and the ground-truth disparities. Conversely, we follow an unsupervised approach: given a left image $\vect{I}_l$ and a right image $\vect{I}_r$, we are interested in predicting the disparity maps $\vect{d}_r$ and $\vect{d}_l$. The disparity map $\vect{d}_r$ is defined as the 2D map where each pixel value represents the offset of the corresponding pixel from the left and the right images. Symmetrically, $\vect{d}_l$ encodes the offsets from the right to the left images.
We propose to estimate the disparity in an indirect way through image synthesis from different views.
Specifically, the approach consists in training a network to predict disparity maps that can be used to generate the left images from the right images or vice-versa. 
Formally speaking, 
we assume that a right-to-left disparity map $\vect{d}_l$ is produced from a generator network $G$ with both the left and right images, $\vect{I}_l$ and $\vect{I}_r$, as inputs. The warping function $f_w(\cdot)$ is used to perform the synthesis of the left image view by sampling from $\vect{I}_r$
\begin{equation}
  \hat{\vect{I}}_l = f_w(\vect{d}_l, \vect{I}_r).\label{eq:samp}
\end{equation}
with
\begin{equation}
  \vect{d}_l,\vect{d}_r = G(\vect{I}_l, \vect{I}_r).\label{eq:net}
\end{equation}
Importantly, the image sampler used to implement the warping function $f_w(\cdot)$ needs to be differentiable in order to be able to train the whole model via gradient descent. Therefore, we use the image sampler from the spatial transformer network \cite{jaderbergNips15} that employs a bilinear sampler.
A reconstruction loss between $\hat{\vect{I}}_l$ and $\vect{I}_l$ is thus utilized to provide supervision in optimizing the network $G$. Usually, the $\mathcal{L}_{1}$ loss is employed:
\begin{equation}
  \mathcal{L}_{rec}^l =\lVert \vect{I}_l -\hat{\vect{I}}_l  \lVert_1.\label{lrecl}
  \end{equation}
Symmetrically, we use the left-to-right disparity $\vect{d}_r$ to synthesize the left image:
\begin{equation}
	\hat{\vect{I}}_r = f_w(\vect{d}_r, \vect{I}_l).
\end{equation}
and obtain the corresponding loss:
\begin{equation}
	\mathcal{L}_{rec}^r =\lVert \vect{I}_r -\hat{\vect{I}}_r  \lVert_1.
\end{equation}
Finally, if we assume that the images are rectified, and that we know the baseline distance $b$ between the two cameras and the focal length $f$, we can obtain
the depth at a pixel location $(x,y)$ of the left image from the predicted disparity with $d_l=b\frac{f}{d(x,y)}$.
We now detail how this general unsupervised approach can be extended to a cycle binocular model.

\subsection{Network Training for Binocular Depth Estimation}
\label{sec:USDE}
In this section, we detail the training loss employed in our binocular depth estimation model. 
The reconstruction loss is defined as the sum of the reconstruction losses of the two images.
\begin{equation}
 \mathcal{L}_{rec} = \mathcal{L}_{rec}^r +\mathcal{L}_{rec}^l\label{lrec}
\end{equation}
where $\mathcal{L}_{rec}$ is defined as in Eqn. \eqref{lrecl}. In order to constrain the predicted disparities on each other, we also add an $L1$-norm consistency loss as follows:
\begin{equation}
\mathcal{L}_{con} = || \vect{d}_l - f_w(\vect{d}_l, \vect{d}_r) ||_1+|| \vect{d}_r - f_w(\vect{d}_r, \vect{d}_l) ||_1\label{lcon}
\end{equation}
Since the two disparity maps correspond to different views, they are not aligned and their consistency cannot be measured directly with an $\mathcal{L}_1$ loss . Inspired by \cite{godard2017unsupervised}, we use the warping operation to make them pixel-to-pixel aligned. More precisely, in the first term of Eqn. \eqref{lcon}, since the left-to-right disparity $\vect{d}_r$ is aligned with the right image, we use the right-to-left warping $f_w(\vect{d}_l, .) $ introduced in Eqn. \eqref{eq:samp} to obtain a disparity aligned with $\vect{d}_l$.

\begin{figure*}[!t]
\begin{center}
\subfloat[The network is composed of two streams that take as input the left and the right images respectively. ]{\includegraphics[width=0.7\textwidth]{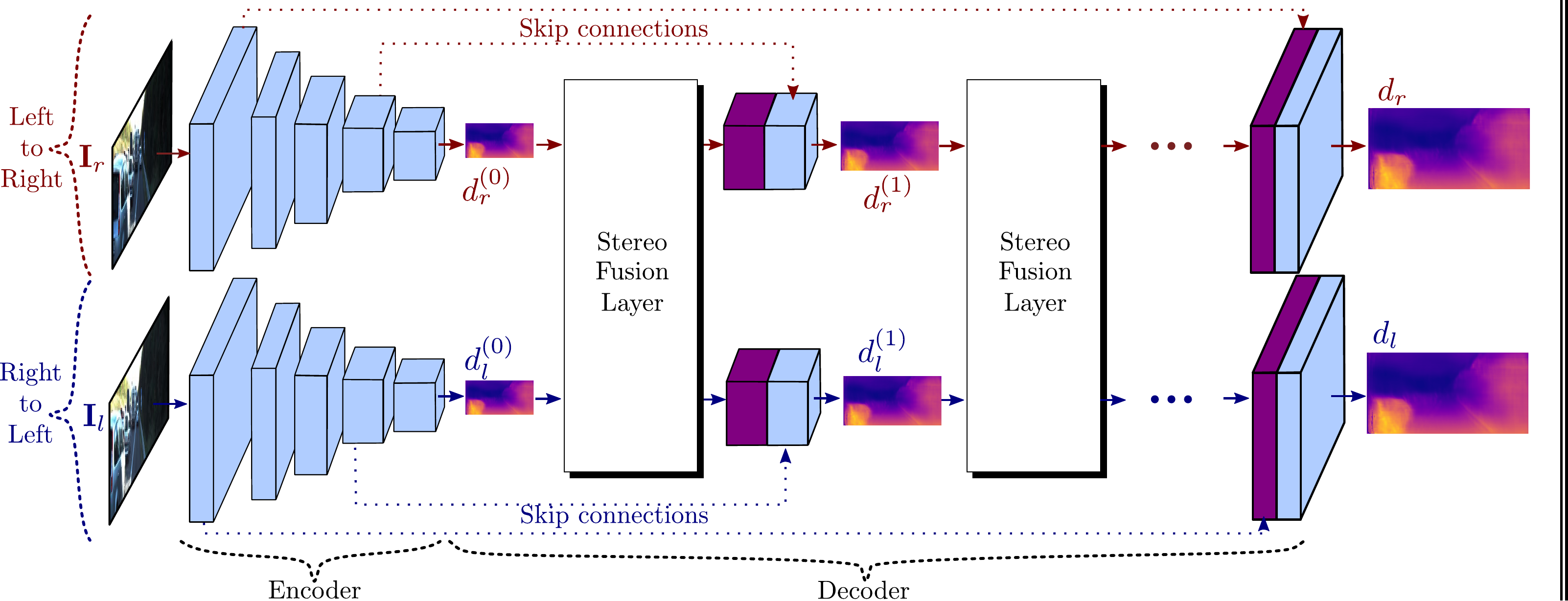}}
\hspace{0.2cm}
\subfloat[Stereo Fusion layer]{\includegraphics[width=0.2\textwidth]{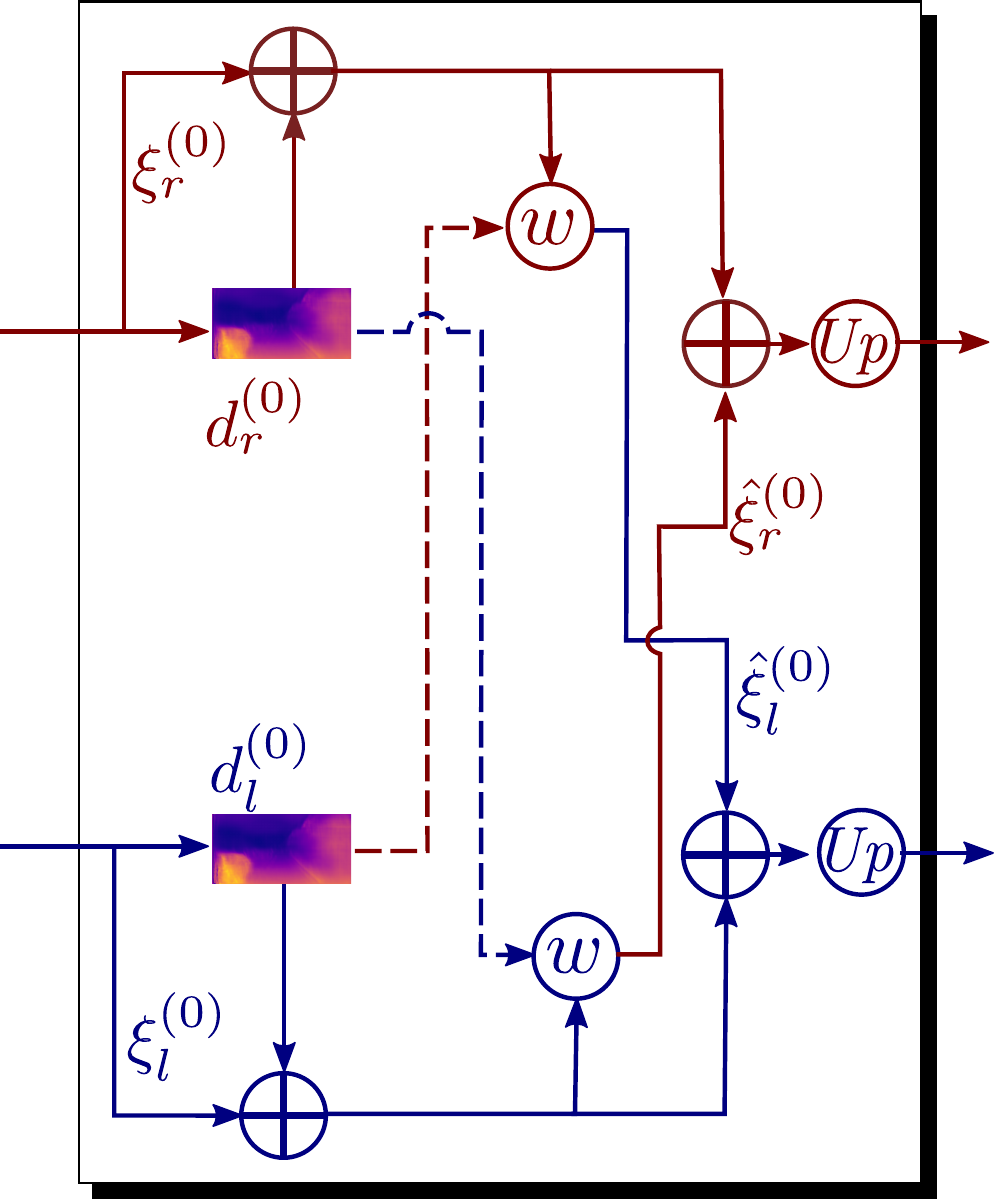}}
\caption{Illustration of the proposed Progressive Fusion Network (PFN). In the left-to-right stream (in dark red) all tensors are aligned with right image. Conversely, in the right-to-left stream (in dark blue) all the tensors are aligned with the left image. The estimated left-to-right disparity $\vect{d}_r^{(0)}$ is used to align the left image feature map $\vect{\xi}_l^{(0)}$ and the right-to-left disparity $\vect{d}_l^{(0)}$ with the left-to-right stream. The aligned tensor $\hat{\vect{\xi}}_r^{(0)}$ is then concatenated with the right-to-left stream. Skip connections (dotted lines) are used to transfer local information from the encoder to the decoder.  $\oplus$ denotes the concatenation operator, \textcircled{\tiny$w$} denotes the warping operator introduced in \eqref{eq:warp}, \textcircled{\tiny Up} denotes the $2\times 2 $ Up-sampling operator.}
\label{fig:funet}
\end{center}

\vspace{-12pt}
\end{figure*}

In order to further improve the quality of the synthesized images, we also propose to use adversarial learning~\cite{goodfellow2014generative}. 
The key idea of adversarial learning is to train two networks simultaneously, a discriminator and a generator. The objective of the generator is to generate realistic images (in our case the right image from the left image and vice-versa). The goal of the discriminator is to distinguish real images of the training set from generated images. In our particular case, we add two discriminators $D_r$ and $D_l$.
The discriminator network $D_r$ is trained to distinguish real right images $\vect{I}_r$ from right images that were synthesized from left images $\hat{\vect{I}}_r$. Similarly, $D_l$ is trained to distinguish real left images $\vect{I}_l$ from left images $\hat{\vect{I}}_l$ that were synthesized from right images. 
In \cite{goodfellow2014generative} the proposed adversarial loss is formulated as a min-max objective function that involves a cross entropy loss.
However, in this standard GAN formulation, the optimization generally suffers from vanishing gradients due to the sigmoid cross-entropy loss. There have been recent improvements in the GAN methodology to stabilize training and, to this aim, we use a least-square GAN loss \cite{mao2017least} by substituting the cross-entropy loss by
the least-squares function with binary coding (1 for real, 0 for synthesized). Consequently, the formulation in Eqn. \eqref{eq:std_gan} is split in two losses used to trained the discriminator and the generator respectively:
\begin{equation}
\begin{split}
  \mathcal{L}_{gan}^{D,r}(D_r) = &
  \mathbb{E}_{\vect{I}_r \sim p(\vect{I}_r)}[(D_r(\vect{I}_r)-1)^2]  \\
  &+  \mathbb{E}_{\vect{I}_l \sim p(\vect{I}_l)}[D_r(f_w(\vect{d}_r, \vect{I}_l))^2]
 \end{split}
\label{eq:ganDisc}
\end{equation}
\begin{equation}
\begin{split}
\mathcal{L}_{gan}^{G,r}(G) = &\mathbb{E}_{\vect{I}_l \sim p(\vect{I}_l)}[(D_r(f_w(\vect{d}_r, \vect{I}_l))-1)^2]
\end{split}
\label{eq:std_gan}
\end{equation}
The intuition behind Eqn. \eqref{eq:ganDisc} is that the discriminator is trained to output~1 when the input image is a real right images and~0~when the input is a synthesized image. In Eqn. \eqref{eq:ganDisc}, the $G$ network is trained in order to predict a $d_r$ disparity map such that the discriminator confuses the synthesized image with real right images and, thus, outputs 1.  The total adversarial loss for the left-to-right stream is given by:
\begin{equation}
\mathcal{L}_{gan}^{r}=\mathcal{L}_{gan}^{D,r}+\mathcal{L}_{gan}^{G,r}\label{lganr}
\end{equation}
We define similarly the adversarial loss for the right-to-left stream $\mathcal{L}_{gan}^{l}$ and obtain the total adversarial:
\begin{equation}
\mathcal{L}_{gan}=\mathcal{L}_{gan}^{r}+\mathcal{L}_{gan}^{l}\label{lgan}
\end{equation}
A major advantage of considering an adversarial loss is that it imposes a global consistency loss oppositely to the $\mathcal{L}_1$ loss used in Eqn.\eqref{lrec} that acts only locally.
Note that, at test time, the inferred $\vect{d}_l$ and $\vect{d}_r$ are used as final outputs of the model and the discriminators are not used anymore. 


\subsection{Cycled Generative Networks for Binocular Depth Estimation}
\label{sec:cycle}

In order to further exploit the left and right images synthesized by our half-cycle network, we propose a cycled network structure. An overview of the proposed framework is shown in Fig.~\ref{fig:detailedgan}. A first generator network, forward half-cycle, produces two distinct disparity maps ($\emph{d}_r$, $\emph{d}_l$) from different view directions, and synthesizes different-view images as described in Section~\ref{sec:USDE}, namely $\hat{\emph{I}_r}$,$\hat{\emph{I}_l}$. In the second half-cycle, the generator network $G$ takes as inputs these two synthesized images $\hat{\emph{I}_r}$,$\hat{\emph{I}_l}$ and predicts new disparity maps $\emph{d}'_r$, $\emph{d}'_l$ that are again used to synthesize the opposite views $\hat{\emph{I}}'_r$,$\hat{\emph{I}}'_l$ from the synthesized images.~The overall model we obtain in this way forms a cycle.  
This cycle formulation can be seen as a data augmentation approach
since, at training time, the network learns to predict disparity maps
from the images of the training (in the forward half-cycle), but also from
synthesized images (in the backward half-cycle). In the literature, standard methods using GANs for data augmentation~\cite{wang2018low}, use generally two separated networks: a data generator network and the network finally used for prediction. In our case, we employ only a single network exploiting the left-right consistency of the data since the $G$ network is used both to generate training data and to estimate depth.
\addnote[consistency]{1}{
The second half-cycle prevents the first half-cycle network from predicting inconsistent disparity pairs. Indeed inconsistencies in the disparities predicted in the forward half-cycle would harm the estimations of the second half-cycle network. Consequently, imposing cycle consistency favors consistent predictions in the first half-cycle. At inference time, the second half-cycle is not used anymore.}
~More formally, the cycled generative network is
based on the half-cycle structure previously described. Assuming that we obtained the synthesized image $\hat{\vect{I}}_r$ and $\hat{\vect{I}}_l$ from the half-cycle network, we now aim at predicting the original left image from $\hat{\vect{I}}_r$ and the original right image from $\hat{\vect{I}}_l$. To this aim, $\hat{\vect{I}}_r$ and $\hat{\vect{I}}_l$ are used as input of the next cycle generative network $G$. $G$ produces again two disparity maps $\vect{d}_l'$ and $\vect{d}_r'$.  Again, we synthesize the left-view image $\hat{\vect{I}}_l'$ from $\hat{\vect{I}}_r$ and $\hat{\vect{I}}_r'$ from $\hat{\vect{I}}_l$ via the warping operation $f_w$.
Similarly to the forward half-cycle, a reconstruction loss $\mathcal{L}_{rec}'$ is used for the backward half-cycle as in Eqn. \eqref{lrec}. We also add a consistency loss $\mathcal{L}_{rec}'$ and an adversarial $\mathcal{L}_{gan}'$ as in Eqn.\eqref{lcon} and Eqn.\eqref{lgan}, respectively. 
During adversarial learning, synthesized and real images are independently passed to the discriminator networks.

The full optimization objective consists on the combination of the reconstruction, adversarial and consistency losses for both half-cycles and can be written as follows:
\begin{align}
  \mathcal{L} =~& \gamma_1 (\mathcal{L}_{rec} +\mathcal{L}_{rec}') + \gamma_2 (\mathcal{L}_{gan}+\mathcal{L}_{gan}')\notag \\ 
  &+ \gamma_3 (\mathcal{L}_{con}+\mathcal{L}_{con}').
\end{align}
where $\{\gamma_i\}_{i=1}^3$ represents a set of weights for controlling the importance of different terms.

Importantly, when the optimization is finished, given a testing pair $\{\vect{I}_l, \vect{I}_r\}$, the testing is performed using only the first half-cycle. Therefore, the proposed cycle approach does not increase the testing computation time but only the training complexity. Nevertheless, note that this cycle framework increases the training complexity by increasing the number of computation operations but it does not increase the number of parameters.
\addnote[share-discr]{1}{Note that the discriminators of the two half-cycles share their parameters in order to avoid an increase in the number of parameters. In other words, we use a right view discriminator for the right images ($\vect{I}_r$, $\hat{\vect{I}}_r$, $\hat{\vect{I}}_r'$) and a left view discriminator for the left view images ($\vect{I}_l$, $\hat{\vect{I}}_l$, $\hat{\vect{I}}_l'$), they are denoted in Fig.~\ref{fig:detailedgan} as $D_r$ and $D_l$ respectively.}

\subsection{Progressive Fusion Network} \label{subsec:cfn}

\begin{table*}[t]
\centering
\resizebox{.80\textwidth}{!}{
\begin{tabular}{ |>{}c|>{}c>{}c>{}c|>{}c>{}c|>{}c>{}c|}
\toprule
Layer & K & S & Channels & \multicolumn{2}{>{}c|}{left to right branch} & \multicolumn{2}{>{}c|}{right to left branch}\\
\midrule
\multicolumn{4}{|>{}c|}{Encoder (layers share weights among branches)}& Input & Output & Input & Output\\
\midrule
conv      &7&2& 64&      $I_L$ &                       $conv1_{L2R}$ &      $I_R$                &      $conv1_{R2L}$\\
pool       &3&1& 64&      $conv1_{L2R}$ &      $pool1_{L2R}$ &       $conv1_{R2L}$ &      $pool1_{R2L}$\\
ResNetBlock&3&& 256&   $pool1_{L2R}$ &       $resblock1_{L2R}$& $pool1_{R2L}$   &     $resblock1_{R2L}$\\
ResNetBlock&4&& 512&   $resblock1_{L2R}$& $resblock2_{L2R}$ &$resblock1_{R2L}$ &$resblock2_{R2L}$\\
ResNetBlock&6&& 1024& $resblock2_{L2R}$ &$resblock3_{L2R}$ &$resblock2_{R2L}$ &$resblock3_{R2L}$\\
ResNetBlock&3&& 2048&$resblock3_{L2R}$ & $resblock4_{L2R}$ &$resblock3_{R2L}$&$resblock4_{R2L}$\\
\midrule
\multicolumn{4}{|>{}c|}{Decoder (layers do not share weights among branches)} & Input & Output & Input & Output\\
\midrule
UpConv   &3&2& 512&  $resblock4_{L2R}$                                & $upconv6_{L2R}$ & $resblock4_{R2L}$                                    & $upconv6_{R2L}$\\
conv    &3&1& 512&  $upconv6_{L2R}$ + $resblock3_{L2R}$ & $iconv6_{L2R}$     & $upconv6_{R2L}$ + $resblock3_{R2L}$ & $iconv6_{R2L}$ \\
UpConv   &3&2& 256&  $iconv6_{L2R}$                                       &$upconv5_{L2R}$&  $iconv6_{R2L}$                                         &$upconv5_{R2L}$\\
conv    &3&1& 256&  $upconv5_{L2R}$ + $resblock2_{L2R}$ & $iconv5_{L2R}$     &$upconv5_{R2L}$ + $resblock2_{R2L}$  &$iconv5_{R2L}$\\
UpConv   &3&2& 128&  $iconv5_{L2R}$                                       & $upconv4_{L2R}$ &$iconv5_{R2L}$                                          &$upconv4_{R2L}$\\
conv    &3&1& 128&  $upconv4_{L2R}$ + $resblock1_{L2R}$ &  $\xi^0_{r}$     &$upconv4_{R2L}$ + $resblock1_{R2L}$   &$\xi^0_{l}$\\
conv      &3&1& 1&        $\xi^0_{r}$                                      & $\emph{d}^0_{r}$      &$\xi^0_{l}$                                &$\emph{d}^0_{l}$\\
UpConv   &3&2& 64&   $\xi^0_{r}$                                      & $upconv3_{L2R}$   &$\xi^0_{l}$                                          &$upconv3_{R2L}$\\
bilinear upsampling  &-&-& 1&         $\emph{d}^0_{r}$        & $up\text{-}\emph{d}^0_{r}$   &$\emph{d}^0_{l}$                                   &$up\text{-}\emph{d}^0_{l}$\\
iconv3    &3&1& 64&   $upconv3_{L2R}$ + $pool1_{L2R}$ + $up\text{-}\emph{d}^0_{r}$ + $\hat{\xi}^0_l$ & $\xi^1_r$ & $upconv3_{R2L}$ + $pool1_{R2L}$ + $up\text{-}\emph{d}^0_{l}$ + $\hat{\xi}^0_r$ & $\xi^1_l$\\
disp3     &3&1& 1&    $\xi^1_{r}$                                      & $\emph{d}^1_{r}$      &$\xi^1_{l}$                                &$\emph{d}^1_{l}$ \\
upconv2   &3&2& 32&   $\xi^1_{r}$                                      & $upconv2_{L2R}$   &$\xi^1_{l}$                                          &$upconv2_{R2L}$\\
updisp3  &-&-& 1&    $\emph{d}^1_{r}$        & $up\text{-}\emph{d}^1_{r}$   &$\emph{d}^1_{l}$                                   &$up\text{-}\emph{d}^1_{l}$\\
iconv2    &3&1& 32&   $upconv2_{L2R}$ + $conv1_{L2R}$ + $up\text{-}\emph{d}^1_{r}$ + $\hat{\xi}^1_l$ & $\xi^2_r$ & $upconv2_{R2L}$ + $conv1_{R2L}$ + $up\text{-}\emph{d}^1_{l}$ + $\hat{\xi}^1_r$ & $\xi^2_l$\\
disp2     &3&1& 1&   $\xi^2_{r}$                                      & $\emph{d}^2_{r}$      &$\xi^2_{l}$                                &$\emph{d}^2_{l}$ \\
upconv1   &3&2& 16&   $\xi^2_{r}$                                      & $upconv1_{L2R}$   &$\xi^2_{l}$                                          &$upconv1_{R2L}$\\
updisp2  &-&-& 1&    $\emph{d}^2_{r}$        & $up\text{-}\emph{d}^2_{r}$   &$\emph{d}^2_{l}$                                   &$up\text{-}\emph{d}^2_{l}$\\
iconv1    &3&1& 16&   $upconv1_{L2R}$ + $up\text{-}\emph{d}^2_{r}$ + $\hat{\xi}^2_l$ & $\xi^3_r$ & $upconv1_{R2L}$ + $up\text{-}\emph{d}^2_{l}$ + $\hat{\xi}^2_r$ & $\xi^3_l$\\
disp1     &3&1& 1&    $\xi^3_{r}$                                      & $\emph{d}^3_{r}$      &$\xi^3_{l}$                                &$\emph{d}^3_{l}$ \\
\bottomrule
\end{tabular}
}
\vspace{-4pt}
\caption{Detailed architecture of the proposed network, for readability reasons we show the half-cycle structure. $K$, $S$ and $Channels$ denote convolutions kernel size, stride and output channels respectively. For ResNet blocks $K$ denotes the number of blocks. The $+$ indicates concatenation between feature maps.}
\label{tab:detailed_archi}
\vspace{-10pt}
\end{table*}

 \label{sec:CFN}
  When it comes to binocular depth estimation, the question of how to fuse
the information provided by each image needs to be addressed. A
standard approach, as used in \cite{godard2017unsupervised}, consists in simply
concatenating the two images over the color axis. We denote this
approach as \emph{early fusion}. On the contrary, in our previous work \cite{pilzer2018unsupervised}, we
used a \emph{late fusion} approach that consists in estimating the two
disparities separately employing two separated networks, before fusing them. In this section, we
first explain why these two approaches suffer from the misalignment between the input
images and the output disparity map. Secondly, we propose a neural network
architecture to face this issue. Let us consider again the case
in which we aim at estimating the right-to-left disparity
$\vect{d}_l$ from the images $\vect{I}_l$ and $\vect{I}_r$. When
looking at the images, we can notice that the edges in
$\vect{I}_l$ are perfectly aligned with the edges of
$\vect{d}_l$. This observation results directly from the disparity definition. Therefore, in order to estimate
$\vect{d}_l$ at a pixel location $(u,v)$, the model needs to
look at the pixel values of $\vect{I}_l$ in the neighbour of the pixel
$\vect{I}_l(u,v)$. Conversely, the edges in  $\vect{d}_l$ and $\vect{I}_r$ are not aligned. More precisely, the model would need
to look at the pixel around $\vect{I}_r(u+\vect{d}_l(u,v),v)$ in
order to estimate $\vect{d}_l(u,v)$.

In the context of convolutional neural network, this observation leads
to two conclusions. First, when using an early fusion approach, the
local information can be fused by the network, in practice, only after several layers,
when the receptive field of the activations are larger than the disparity value
we want to estimate. Second, in the case of a late fusion network, the
$d_l$ disparity estimated from $\vect{I}_l$ will have better edges
since the network can benefit from the alignment. Conversely, $\vect{I}_r$ will have a lower quality since the corresponding network
has to handle the input-output misalignement. Therefore, the
benefit brought by the use of $\vect{I}_r$ will not be substantial.

To tackle this issue, we propose a Progressive Fusion Network (PFN). The key
idea behind PFN is to first estimate low resolution disparity maps
that are then used to align the image features. These aligned feature
maps are employed to refine the disparity maps at the higher resolutions. This method is applied iteratively until we obtain
the desired high resolution disparity maps.
This iterative procedure is well in line with the multi-scale
approaches that have shown good performances in the monocular supervised
setting (see Sec~\ref{sec:related}). The details of the
architecture are given in Fig. \ref{fig:funet}.

We first apply an encoder network on each input images obtaining two feature maps $\vect{\xi}_r^{(0)}$ and $\vect{\xi}_l^{(0)}$. 
In our particular case, we use a ResNet-50 architecture since it has already shown good performances on the depth estimation problem \cite{laina2016deeper,godard2017unsupervised}.
We then estimate the left and right low resolution disparities ($\vect{d}_l^{(0)}$ and $\vect{d}_r^{(0)}$ respectively) from
$\vect{\xi}_l^{(0)}$ and $\vect{\xi}_r^{(0)}$ respectively. To do so, we employ a single $3\times 3 $ convolutional layer with sigmoid activations. Now that we have a first estimation of the disparity from the left to the right image, we can employ this disparity $\vect{d}_r^{(0)}$ to warp the feature maps $\vect{\xi}_l^{(0)}$ and the disparity $\vect{d}_l^{(0)}$ from the opposite stream:
\begin{equation}
\hat{\vect{\xi}}^{(0)}_r = f_w(\vect{d}^{(0)}_r, \vect{\xi}^{(0)}_l\oplus\vect{d}^{(0)}_l)\label{eq:warp}
\end{equation}
where $\oplus$ denotes the concatenation operator. By concatenating the features and the disparity, we provide to the left-to-right stream all the information currently available in the right-to-left stream. We obtain a complete left image representation that is aligned with the right image. Symmetrically,  $\vect{d}_l^{(0)}$ is used to warp the feature maps $\vect{\xi}_r^{(0)}$ and $\vect{d}_l^{(0)}$ computed in the opposite stream according to $ \hat{\vect{\xi}}^{(0)}_l = f_w(\vect{d}_l^{(0)}, \vect{\xi}^{(0)}_r\oplus\vect{d}^{(0)}_r)$.
    Then, we
concatenate  $\vect{\xi}_r^{(0)}$, $\hat{\vect{\xi}}_r^{(0)}$ and
$\vect{d}_r^{(0)}$ and perform $2\times 2$ up-sampling. Finally, the next
resolution feature map $\vect{\xi}_r^{(1)}$ is obtained by concatenation with the feature maps of the encoder with the same dimension as in a standard U-Net \cite{ronneberger2015u}. The skip connections are employed to transfer directly the local information from the encoder to the decoder.
A similar operation is
applied on the left network leading to $\vect{\xi}_l^{(1)}$. All these
operations are performed four times in order to obtain the full
resolution disparities.

In order to further benefit from the multi-scale approach, we employ the $L1$-norm reconstruction loss
$\mathcal{L}_{rec}$ at every resolution $i \in \{0..3\}$ for both the right and left images:
\begin{align}
 \mathcal{L}_{rec}^{(i)} = \lVert \vect{I}^{(i)}_l - f_w(\vect{d}^{(i)}_l,
 \vect{I}^{(i)}_r) \lVert_1 \notag 
+ \lVert \vect{I}^{(i)}_r - f_w(\vect{d}^{(i)}_r, \vect{I}^{(i)}_l).
 \lVert_1, 
 \end{align}
Note that $\mathcal{L}_{rec}^{(3)}$ corresponds to the highest dimension, and in this way, to the loss given in Eqn. \eqref{lrec}. Consequently, when training this multi-scale model, Eqn. \eqref{lrec} is replaced by the following multi-scale loss:
\begin{equation}
\mathcal{L}_{rec}=\sum_{i=0}^3 \mathcal{L}_{rec}^{(i)}.
\end{equation}

A multi-scale loss is also employed in \cite{godard2017unsupervised,garg2016unsupervised}, however, in our model, the low resolution depth maps are not only used to deeply supervise the network as in \cite{godard2017unsupervised}. Instead, the low resolutions depth maps are used to correct the misalignment between the images and, in this way, help the network to better predict the higher resolutions.

\subsection{Network Implementation Details}
We now describe the details of the network implementation.
For the encoder of $G$, we use a ResNet-50
backbone network as in \cite{laina2016deeper}. The left and right
encoders share the weights. Conversely, the forward and the
backward cycle paths share their parameters. 
For the discriminators $D_l$ and $D_r$, we employ a network
structure which has five consecutive convolutional operations with a
kernel size of 3, a stride size of 2 and a padding size of 1, and
batch normalization~\cite{ioffe2015batch} is performed after each
convolutional operation. The adversarial loss is applied to output
patches and is implemented following\cite{mao2017least}. 
The encoder network takes as input images of size $256 \times 512$. The ResNet-50 encoder outputs high level features of size $(4 \times 8 \times 2048)$. As mentioned in Sec~\ref{sec:CFN}, each up-sampling is followed by a convolution layer. We employ $3\times 3$ convolution layers with number of channels of 512, 256, 128 and 64 respectively 
 with \emph{Elu} activations. 

The detailed architecture of our network is described in Table~\ref{tab:detailed_archi} where we present the structure of our half-cycle network. We specify the inputs and outputs of each part of the network. In particular, following the notation of previous Section~\ref{sec:CFN} we denoted with $\hat{\xi}$ the concatenation of features and disparity of one branch of the network after the warping that aligns them with the other branch.

\section{Experimental Results}
\label{sec:exp}

\begin{table*}[!t]
\begin{center}
\resizebox{.75\textwidth}{!}{
\begin{tabular}{ | l | c || c | c | c | c | c | c | c |}
\toprule
   \multirow{2}{*}{Method} & \multirow{2}{*}{Warping} & Abs Rel & Sq Rel & RMSE & RMSE log & $\delta<1.25$ & $\delta<1.25^2$ & $\delta<1.25^3$ \\
   \cline{3-6} \cline{7-9}
    && \multicolumn{4}{c|}{lower is better}& \multicolumn{3}{c|}{higher is better} \\
\midrule
\textit{Half-Cycle Mono}+ ${L}_{gan}$\cite{pilzer2018unsupervised}& \emph{w/o} & 0.165 & 1.756 & 6.164 & 0.257 & 0.773 & 0.914 & 0.962 \\
\textit{Half-Cycle Stereo} + ${L}_{gan}$\cite{pilzer2018unsupervised}&\emph{w/o}&0.163 &1.620 &6.129 &0.254 &0.770 &0.913 &0.962 \\
\textit{Cycle Stereo} + ${L}_{gan}$\cite{pilzer2018unsupervised} &\emph{w/o}&0.153 &1.388 &6.016 &0.247 &0.789 &0.918 &0.965 \\
\midrule
\textit{Half-Cycle Stereo}   &$d$ &0.159 &1.374 &6.105 &0.261 &0.764 &0.909 &0.960\\      
\textit{Half-Cycle Stereo} &$d\oplus\xi$ &0.153 &1.260 &5.960 &0.254 &0.777 &0.915 &0.963\\
\textit{Half-Cycle Stereo} + ${L}_{gan}$ &$d\oplus\xi$ &0.148 &1.209 &5.827 &0.246 &0.789 &0.921 &0.966 \\
\midrule

\textit{Cycle Stereo}  &$d$&0.146 &1.246 &5.833 &0.239 &0.791 &0.922 &0.968\\
\textit{Cycle Stereo}  &$d\oplus\xi$ &0.141 &1.235 &5.661 &0.234 &0.807 &0.930 &0.970 \\
\textit{Cycle Stereo} + ${L}_{gan}$ &$d\oplus\xi$ &0.137 &1.199 &5.721 &0.234 &0.806 &0.928 &0.970\\
\textit{Cycle Stereo} + ${L}_{gan}$ + SSIM &$d\oplus\xi$ &\bf 0.102 &\bf 0.802 &\bf 4.657 &\bf 0.196 & \bf 0.882 & \bf 0.953 &\bf 0.977\\
\bottomrule
\end{tabular}
}
\end{center}
\vspace{-10pt}
\caption{Quantitative evaluation results of different variants of the proposed approach on the KITTI dataset for the ablation study. 
The estimated depth range is from 0 to 80 meters. $\mathcal{L}_{gan}$ denotes the use of the adversarial loss}
\label{tab:ablation_kitti}
\vspace{-4pt}
\end{table*}

\begin{table*}[h]
\begin{center}
\resizebox{.75\textwidth}{!}{
\begin{tabular}{ | l | c || c | c | c | c | c | c | c |}
\toprule
   \multirow{2}{*}{Method} & \multirow{2}{*}{Warping} & Abs Rel & Sq Rel & RMSE & RMSE log & $\delta<1.25$ & $\delta<1.25^2$ & $\delta<1.25^3$ \\
   \cline{3-6} \cline{7-9}
    && \multicolumn{4}{c|}{lower is better}& \multicolumn{3}{c|}{higher is better} \\
\midrule
    ~\textit{Half-Cycle Mono} \cite{pilzer2018unsupervised}          & \emph{w/o} & 0.468 & 7.399 & 5.741 & 0.493 & 0.735 & 0.890 & 0.945 \\
    ~\textit{Half-Cycle Stereo}  \cite{pilzer2018unsupervised}           & \emph{w/o} & 0.462 & 6.098 & 5.740 & 0.377 & 0.708 & 0.873 & 0.937 \\
    ~\textit{Half-Cycle} + $\mathcal{L}_{gan}$   \cite{pilzer2018unsupervised}         & \emph{w/o} & 0.439 & 5.714 & 5.745 & 0.400 & 0.711 & 0.877 & 0.940 \\
    ~\textit{Cycle} + $\mathcal{L}_{gan}$  \cite{pilzer2018unsupervised}       & \emph{w/o} & 0.440 & 6.037 & 5.443 & 0.398 & 0.730 & 0.887 & 0.944 \\
\midrule
    ~\textit{Half-Cycle Stereo}  & $d$  &0.465 &6.783 &5.503 &0.429 &0.732 &0.887 &0.945 \\
    ~\textit{Half-Cycle Stereo} & $d\oplus\xi$ &0.436 &6.357 &4.877 &0.364 &0.778 &0.915 &0.958 \\
    ~\textit{Half-Cycle Stereo} +$\mathcal{L}_{gan}$ & $d\oplus\xi$ &0.429 &6.304 &5.051 &0.343 &0.778 &0.913 &0.957 \\
\midrule
    ~\textit{Cycle Stereo}  & $d$ &0.445 &6.008 &5.372 &0.488 &0.743 &0.893 &0.947 \\
    ~\textit{Cycle Stereo}  & $d\oplus\xi$ &0.420 &5.767 &4.749 &0.379 &0.790 &0.919 &0.959 \\
    ~\textit{Cycle Stereo} + $\mathcal{L}_{gan}$ & $d\oplus\xi$ &0.418 &5.799 &4.698 &0.343 &0.787 &0.917 &0.959 \\
    ~\textit{Cycle Stereo} + $\mathcal{L}_{gan}$+ SSIM & $d\oplus\xi$  &\bf 0.404 & \bf5.677 & \bf 4.534 & \bf 0.324 & \bf 0.792 & \bf 0.922 & \bf0.962 \\
\bottomrule
\end{tabular}
}
\end{center}
\vspace{-10pt}
\caption{Quantitative evaluation results of different variants of the proposed approach on the Cityscapes dataset for the ablation study. $\mathcal{L}_{gan}$ denotes the use of the adversarial loss.}
\label{tab:ablation_city}
\vspace{-7pt}
\end{table*}

\begin{figure*}
\centering
\includegraphics[width=.7\textwidth]{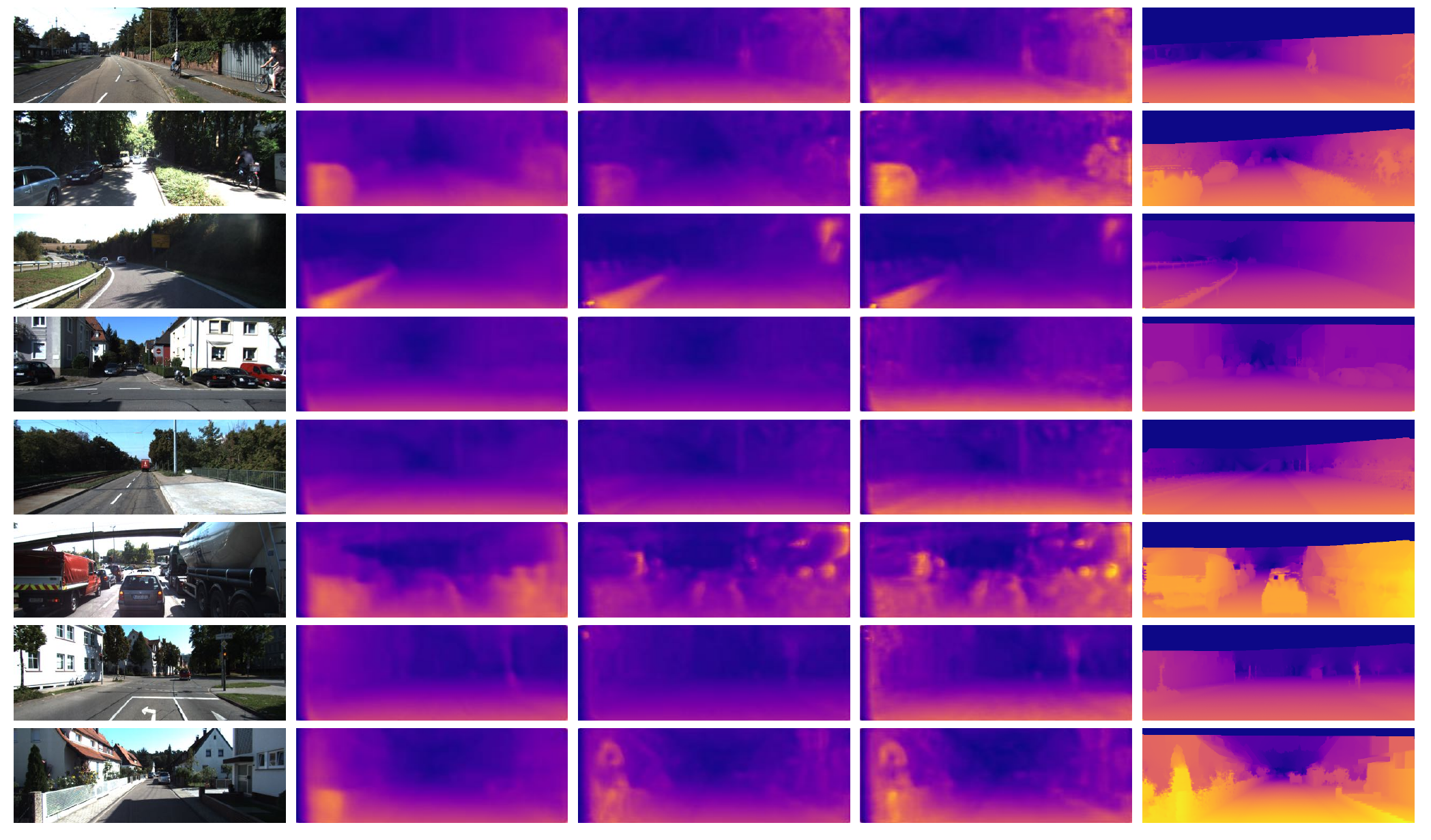}
    \put(-345,214){\scriptsize RGB Image}
	\put(-285,214){\scriptsize\textit{Half-Cycle}, $d$}
	\put(-215,214){\scriptsize\textit{Half-Cycle}, $d\oplus\xi$}
	\put(-150,214){\scriptsize\textit{Half-Cycle} + $\mathcal{L}_{GAN}$, $d\oplus\xi$}
	\put(-55,214){\scriptsize GT Depth Map}
\vspace{-10pt}
\caption{Qualitative comparison of different baseline models of the proposed \textit{Half-Cycle} approach on KITTI Eigen test split. From left to right our stereo disparity progressive fusion, then stereo disparity and features progressive fusion and in the fourth column  the full \textit{Half-Cycle} model with adversarial learning. First column on the left is the RGB images and right the ground truth depth.}
\label{fig:kitti_half}
\vspace{-7pt}
\end{figure*}

\begin{figure*}
\centering
\includegraphics[width=.7\textwidth]{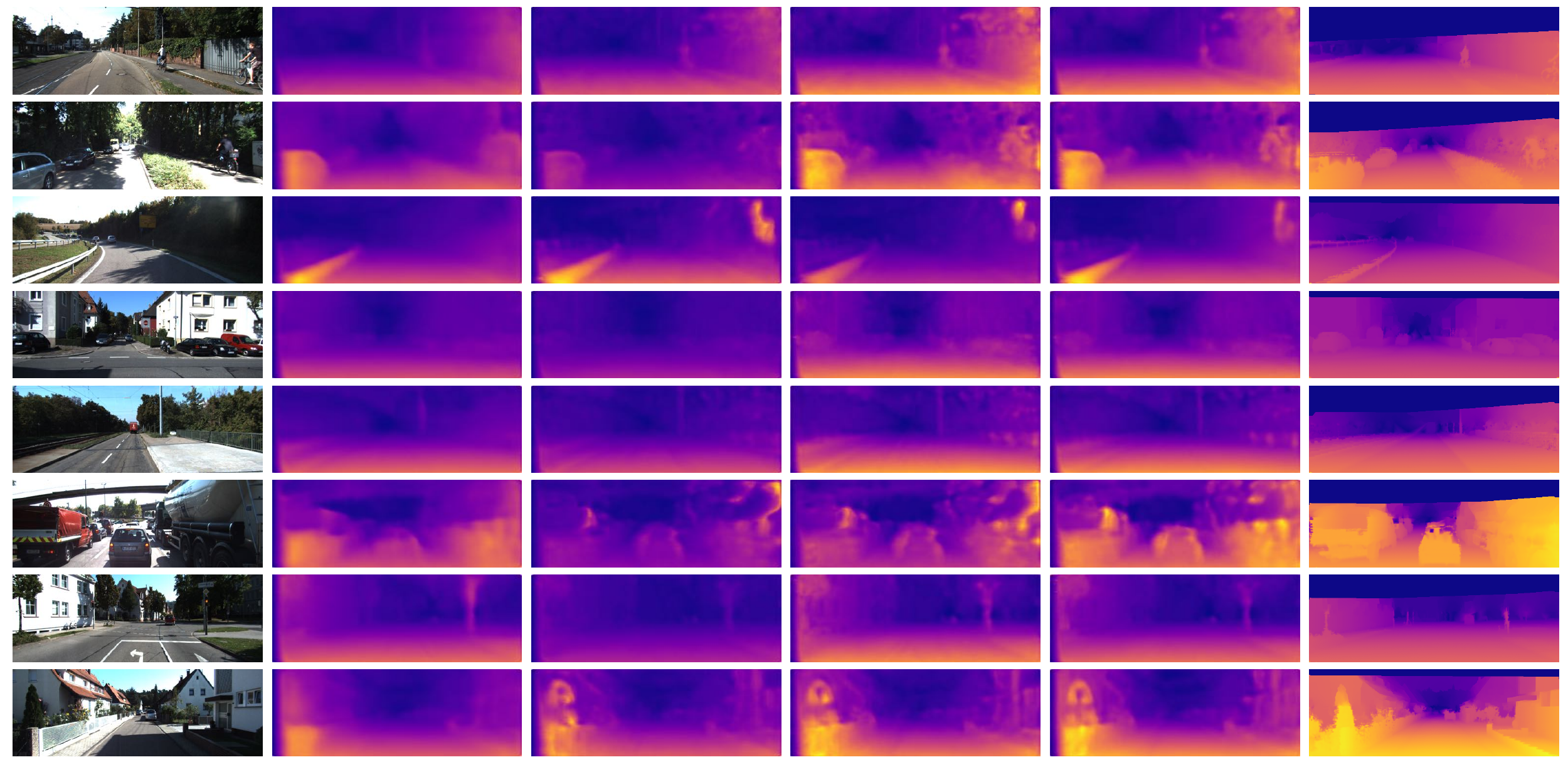}
    \put(-355,180){\scriptsize RGB Image}
	\put(-290,180){\scriptsize \textit{Cycle}, $d$}
	\put(-240,180){\scriptsize \textit{Cycle}, $d\oplus\xi$}
	\put(-195,180){\scriptsize \textit{Cycle} + $\mathcal{L}_{GAN}$, $d\oplus\xi$}
	\put(-125,180){\scriptsize \textit{Cycle} + $\mathcal{L}_{GAN \&SSIM}$}
	\put(-50,180){\scriptsize GT Depth Map}
\vspace{-10pt}
\caption{Qualitative comparison of different baseline models of the proposed \textit{Cycle Stereo} approach on KITTI Eigen test split. From left to right RGB images, \textit{Cycle Stereo} with continuous disparity fusion, \textit{Cycle Stereo} with disparity and features fusion, in column four the full model trained with adversarial learning, in column five the futher refined full model with SSIM (self-similarity) loss and in the right column the ground truth depth maps.}
\label{fig:kitti_cycle}
\vspace{-10pt}
\end{figure*}


\subsection{Experimental Setup}
\textbf{Datasets.} We carry out experiments on three large stereo images datasets, \ie~KITTI~\cite{kitti}, Cityscapes~\cite{Cityscapes} and ApolloScape~\cite{apollo}. For the \textbf{KITTI} dataset, we use the Eigen split~\cite{eigen2014depth} for training and testing. This split contains 22,600 training image pairs, and 697 test pairs. We do data augmentation with online random flipping of the images during training. The \textbf{Cityscapes} dataset is collected using a stereo camera from a driving vehicle through several German cities, during different times of the day and seasons. It presents higher resolution images and is annotated mainly for semantic segmentation. To train our model we combine the densely and coarse annotated splits to obtain 22,973 image-pairs. For testing we use the 1,525 image-pairs of the densely annotated split. The test set also has pre-computed disparity maps for the evaluation. 
The \textbf{ApolloScape} dataset has been collected from a stereo camera attached to a car driving in different Chinese cities. To the best of our knowledge, we are the first to benchmark depth estimation methods on the ApolloScape dataset. We employ two sequences from the \textit{Scene Parsing} data split, scene \textit{road02} and \textit{road03}, obtaining $9156$ training image pairs and $2186$ testing image pairs. Note that, the other sequences use varying setting stereo camera settings and, as a consequence, cannot be used easily for depth estimation. The dataset provides dense depth ground-truth for all the images. 

\vspace{10pt}
\noindent
\textbf{Training Procedure and Parameter Setup.} We train the models denoted with \textit{Half-Cycle Stereo} with a standard training procedure, \ie~initializing the network with random weights and making the network train for 10 epochs. This corresponds to $\approx28K$ steps for both the KITTI and Cityscapes datasets and $11.5K$ steps for ApolloScape. The models denoted with \textit{Cycle Stereo} are optimized starting from the corresponding pre-trained half-cycle model. We train the full cycle model for 10 additional epochs with the same optimization hyper-parameters. 
We use the Adam optimizer for the optimization. The momentum parameter and the weight decay are set to $0.9$ and $0.0002$, respectively. The final optimization objective has weighed loss parameters $\gamma_1=1$, $\gamma_2=0.1$ and $\gamma_3=0.1$. 
The batch size for training is set to 8 stereo image pairs and the learning rate is $10^{-5}$ in all the experiments. Unlike in~\cite{pilzer2018unsupervised} where the learning rate is reduced, in this work it is constant and each experiment is performed for $10$ epochs, significantly reducing the training time. In addition, in~\cite{pilzer2018unsupervised}, the network is trained for 50 epochs, while the model proposed in this work requires only 20 epochs to converge. The simpler training procedure can be explained by the lower number of parameters of our proposed model. Indeed the four decoders used in \cite{pilzer2018unsupervised} are replaced by two decoders that share parameters. \addnote[schedule]{1}{Importantly, in our preliminary experiments, we observed that the SSIM loss is really sensitive to the training schedule (number of iterations and learning decay) on KITTI. In order to draw a fair comparison with \cite{godard2017unsupervised}, we employ the training schedule of \cite{godard2017unsupervised}, when using SSIM on KITTI.}

With respect to the time aspect, the training of the \textit{Half-Cycle Stereo} network models, on two Titan Xp GPUs and KITTI dataset for 10 epochs, takes around $4.5$ hours for the simpler to $7$ hours for the more complex model. The full model \textit{Cycle Stereo} requires 10 additional epochs of training that take from $5$ to $8$ hours depending on the complexity of the model. \addnote[inference-time]{1}{Regarding testing, in our experiments the inference time for each stereo pair is $45$ ms.}

The proposed model is implemented using the deep learning library~\textit{TensorFlow}~\cite{tensorflow2015-whitepaper}. The input images are down-sampled to a resolution of $512 \times 256$ from $1226 \times 370$ in the case of the KITTI dataset, while for the Cityscapes dataset, at the bottom one fifth of the image is cropped following~\cite{godard2017unsupervised} and then is resized to $512 \times 256$. The resolution of the ApolloScape original images is $3384 \times 2710$ pixels. After rectification and cropping using the API of \cite{apollo}, we obtain $2048 \times 1268$ pixel images. We then adopt the standard preprocessing used for the Cityscapes dataset.


\vspace{10pt}
\noindent
\textbf{Evaluation Metrics.} To quantitatively evaluate the proposed approach, we follow several standard evaluation metrics used in previous works~\cite{eigen2014depth, godard2017unsupervised, wang2015towards}. Given $P$ the total number of pixels in the test set and $\hat{d}_i$, $d_i$ the estimated depth and ground truth depth values for pixel $i$, we employ the following metrics: 
\label{sec:metrics}
\begin{enumerate}
    \item 
 Mean relative error (abs rel): 
$\frac{1}{P} \sum_{i=1}^{P} \frac{\parallel \hat{d}_i - d_i \parallel}{d_i}$, 
\item Squared relative error (sq rel): 
$\frac{1}{P} \sum_{i=1}^{P} \frac{\parallel \hat{d}_i - d_i \parallel^2}{d_i}$, 
\item Root mean squared error (rmse): 
$\sqrt{\frac{1}{P}\sum_{i=1}^P(\hat{d}_i - d_i)^2}$, 
\item Mean $\log10$ error (rmse log):
$\sqrt{\frac{1}{P} \sum_{i=1}^{P} \parallel \log \hat{d}_i - \log d_i \parallel^2}$
\item Accuracy with threshold $\tau$, \ie the percentage of $\hat{d}_i$ such that $\delta = \max (\frac{d_i}{\hat{d}_i},\frac{\hat{d}_i}{d_i}) < \alpha^\tau$. If not specified otherwise, we employ $\alpha = 1.25$ and $\tau \in [1,2,3]$ following the procedure used in \cite{eigen2014depth}.
\end{enumerate}

\begin{figure*}[t]
  \centering
    \subfloat[KITTI dataset]{\includegraphics[width=0.6\columnwidth]{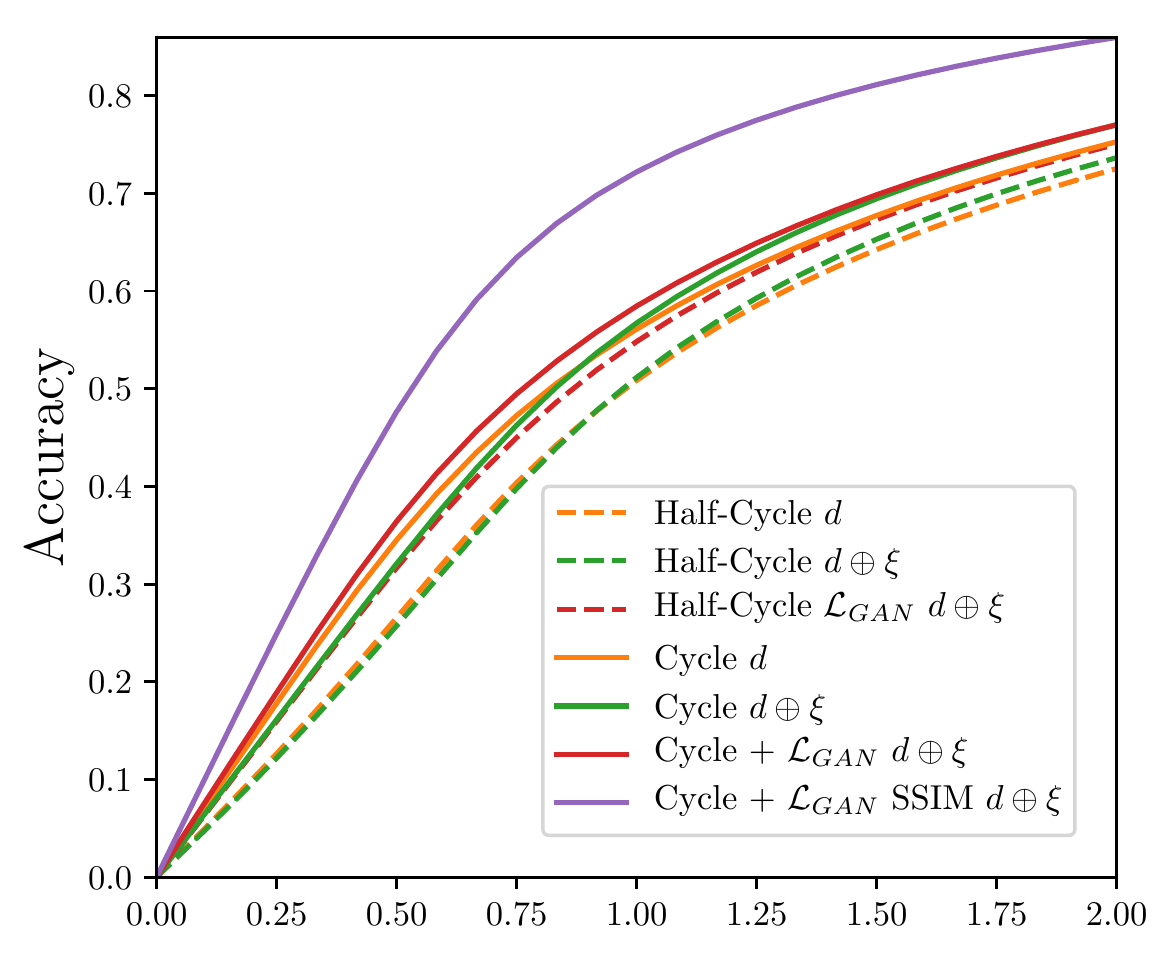}}
    \hspace{-0.3cm}
    \subfloat[Cityscape dataset]{\includegraphics[width=0.6\columnwidth]{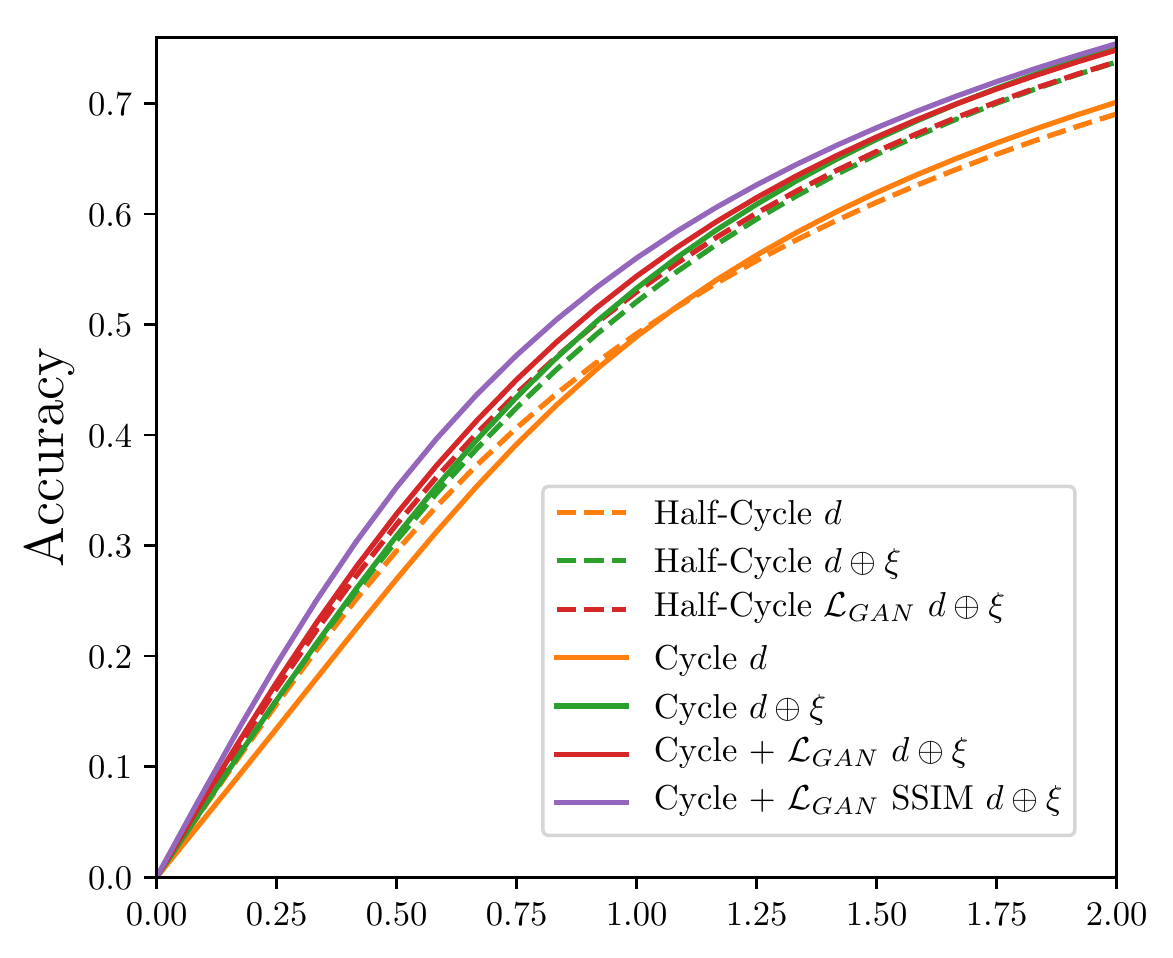}}
    \hspace{-0.3cm}
    \subfloat[ApolloScape dataset]{\includegraphics[width=0.6\columnwidth]{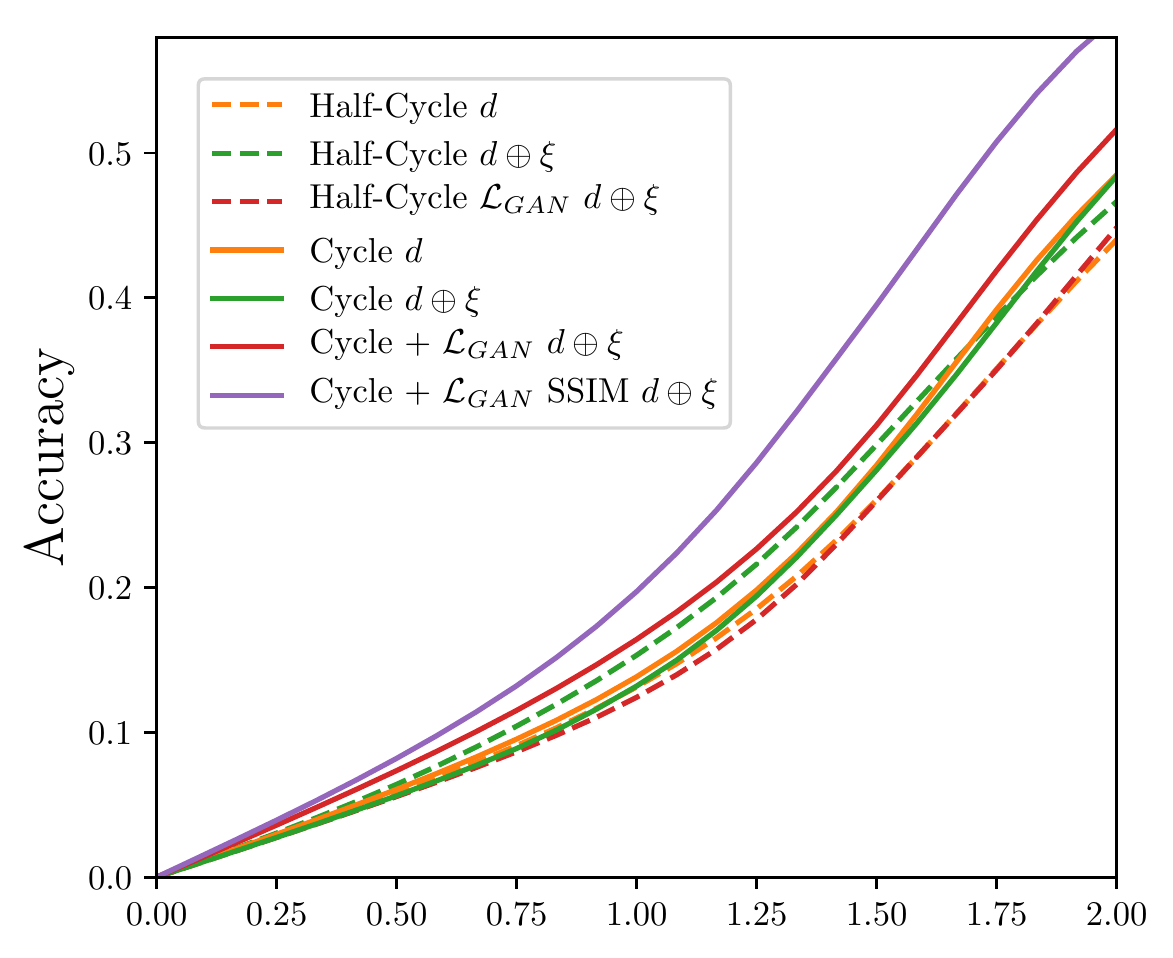}}
    \vspace{-5pt}
    \caption{Accuracy plot with a varying threshold parameter value $\tau$ for the KITTI, Cityscapes and ApolloScape datasets. The accuracy threshold $\alpha$ is set to $\alpha=1.10$ for better visualization}
    \label{fig:plots}
\end{figure*}

\subsection{Ablation study: Baseline Models}

We compare several baseline models for the ablation study:
\begin{enumerate}
\item Half-cycle with a monocular setting (\emph{Half-Cycle Mono}), which uses the forward branch to synthesize from one image view to the other with a single disparity map output and the single RGB image is as input during testing;
\item
  Half-cycle with a stereo setting (\emph{Half-Cycle Stereo}), which uses the forward branch but the network takes as input the two images. It corresponds to the model as described in Sec~\ref{sec:USDE} where G is a PFN as described in Sec~\ref{sec:USDE};
  \item Cycle Stereo, which corresponds to the model as described in Sec~\ref{sec:cycle} where G is also a PFN as described in Sec~\ref{sec:USDE}.
\end{enumerate} We propose to evaluate each model with and without the use of the adversarial loss.
In addition, in order to understand the role of our PFN, we propose to compare three variants of the compared models:
\begin{enumerate}
\item The model without warping (referred to as \emph{w/o} in Tables \ref{tab:ablation_kitti}). In that case, we adopt a late fusion approach as in \cite{pilzer2018unsupervised};
\item A model in which only the estimated disparities (referred to as $d$ in Tables \ref{tab:ablation_kitti} and \ref{tab:ablation_city}) are shuttled to the other stream. Formally speaking, Eqn. \eqref{eq:warp} is replaced by $\hat{\vect{\xi}}^{(0)}_r = f_w(\vect{d}^{(0)}_r,\vect{d}^{(0)}_l)$.
  \item The full model in which both the disparities and the feature maps are shuttled (referred to as $d\oplus\xi$ in  Tables \ref{tab:ablation_kitti} and \ref{tab:ablation_city}). 
\end{enumerate}

\vspace{10pt}
\noindent

\begin{table*}[h]
\begin{center}
\resizebox{.75\textwidth}{!}{
 \begin{tabular}{|>{}l|>{}c||>{}c|>{}c|>{}c|>{}c|>{}c|>{}c|>{}c|}
\toprule
   \multirow{2}{*}{Method} & \multirow{2}{*}{Warping} & Abs Rel & Sq Rel & RMSE & RMSE log & $\delta<1.25$ & $\delta<1.25^2$ & $\delta<1.25^3$ \\
   \cline{3-6} \cline{7-9}
    && \multicolumn{4}{>{}c|}{lower is better}& \multicolumn{3}{>{}c|}{higher is better} \\
\midrule
\textit{Half-Cycle}               &$d         $ &0.485 &14.585 &16.098 &0.452 &0.533 &0.802 &0.897\\
\textit{Half-Cycle}               &$d\oplus\xi$ &0.469 &13.443 &15.779 &0.448 &0.568 &0.797 &0.886 \\
\textit{Half-Cycle} + ${L}_{gan}$ &$d\oplus\xi$ &0.446 &12.283 &14.600 &0.432 &0.571 &0.822 &0.903 \\ 
\midrule
\textit{Cycle}               &$d         $ &0.452 &13.079 &14.927 &0.437 &0.573 &0.815 &0.902\\
\textit{Cycle}               &$d\oplus\xi$ & 0.436 &11.699 &14.661 &0.426 &0.595 &0.810 &0.897\\
\textit{Cycle} + ${L}_{gan}$ &$d\oplus\xi$ &0.423 &11.582 &14.415 &0.422 &0.624 &0.824 &0.905\\ 
\textit{Cycle} + ${L}_{gan}$ + SSIM &$d\oplus\xi$ &\bf 0.387 &\bf 10.097 &\bf 13.449 &\bf 0.396 &\bf 0.669 &\bf 0.843 &\bf 0.915\\ 
\bottomrule
\end{tabular}
}
\end{center}
\vspace{-10pt}
\caption{Quantitative evaluation results of different variants of the proposed approach on the ApolloScape dataset.}
\label{tab:ablation_apollo}
\vspace{-4pt}
\end{table*}

\subsection{Ablation study: Results and discussion}

To validate that the proposed cycled generative network helps and that the proposed PFN is effective for the task, we present an extensive ablation study on both the KITTI dataset (see Table~\ref{tab:ablation_kitti}), on the Cityscape dataset (see Table~\ref{tab:ablation_city}) and on the ApolloScape dataset (see Table~\ref{tab:ablation_apollo}).

First, we observe, that the cycle approach consistently outperforms the \textit{Half-Cycle} approach. For instance on KITTI, if we compare the \textit{Half-Cycle Stereo}+$\mathcal{L}_{GAN}$ model with \textit{Cycle Stereo}+$\mathcal{L}_{GAN}$ in which we warp $d\oplus\xi$, we observe a 0.0114 gain according to the Abs Rel metric, that corresponds to a $7.76\%$ improvement. Similar gain can be observed for all the metrics used in the comparison.
Second, we consistently observe a gain when we employ our \emph{PFN} network with respect to the fusion model proposed in \cite{pilzer2018unsupervised} independently of the use of the cycle approach. Again, this boost in performance brought by the \emph{PFN} is observed on both datasets and according to all the metrics employed in this comparison. Interestingly, we notice that, independently of the use of cycle or adversarial loss, warping both the features and the disparities, performs better than warping only the disparities. We observe that, on both datasets, the adversarial loss helps predicting  better depth maps. It confirms that adding a loss that acts globally can be beneficial for depth estimation. Finally, we report the results when we add the self-similarity loss proposed in \cite{godard2017unsupervised} (referred to as +SSIM in Table \ref{tab:ablation_kitti}), Intuitively, the SSIM loss measures how the object structure in the scene, is preserved in the synthesized image, independently of the average luminance
and contrast. For more technical details, please refer to \cite{wang2004image}. We observe that it further improves the results of our proposed model.

In order to further compare the different baselines, we propose to plot the accuracy metric described in Sec.~\ref{sec:metrics} for different threshold values. More precisely, considering that $\hat{d}_i$, $d_i$ are the estimated and ground truth depth values for pixel $i$, we measure the percentage of $\hat{d}_i$ such that $\delta = \max (\frac{d_i}{\hat{d}_i},\frac{\hat{d}_i}{d_i}) < a^\tau$ when $\tau$ varies. Note that contrary to the scores reported in Table \ref{tab:ablation_kitti} and \ref{tab:ablation_city}, we chose $\alpha=1.1$ for the sake of better visualization. The obtained plots are reported in Fig. \ref{fig:plots}.

\begin{table*}[!t]
\begin{center}
\resizebox{.8\textwidth}{!}{
\begin{tabular}{ | >{}l | >{}c | >{}c | >{}c || >{}c | >{}c | >{}c | >{}c || >{}c | >{}c | >{}c |>{}c |}
\toprule
\multirow{2}{*}{Method} & \multirow{2}{*}{Cycle Inputs} & \multirow{2}{*}{Warping} & \multirow{2}{*}{Dataset}& \multicolumn{4}{>{}c||}{KITTI}& \multicolumn{4}{>{}c|}{ApolloScape} \\
   \cline{5-8} \cline{9-12}

    &&& & Abs Rel & Sq Rel & RMSE & RMSE log & Abs Rel & Sq Rel & RMSE & RMSE log \\
\midrule
\textit{Half-Cycle} + ${L}_{gan}$& - &$d\oplus\xi$ & K &0.148 &1.209 &5.827 &0.246 &0.446 &12.283 &14.600 &0.432\\
\midrule

\textit{Cycle} + ${L}_{gan}$& ($\hat{I}_r$,$I_l$)&$d\oplus\xi$ & K &0.146 &1.466 &5.918 &0.244 &0.441 &12.292 &14.877 &0.438\\
\textit{Cycle} + ${L}_{gan}$& ($\hat{I}_r$,$\hat{I}_l$)&$d\oplus\xi$ & K &\bf 0.137 &\bf 1.199 &\bf 5.721 &\bf 0.234 &\bf 0.423 & \bf 11.582 &\bf 14.415 &\bf 0.422 \\
\bottomrule
\end{tabular}
}
\end{center}
\vspace{-10pt}
\caption{Ablation study: exploiting resynthesized images. We compare two different approaches for exploiting the resynthesized images in our cycle network. We present results for the KITTI dataset (left) and for the ApolloScape dataset (right).}
\label{tab:ablation_reg_kitti}
\vspace{-5pt}
\end{table*}

  On the KITTI dataset, we first notice that the results in the plot are in line with those presented in Table \ref{tab:ablation_kitti} since we clearly observe the benefit of the use of both our cycle setting and the proposed\comRev{~\emph{PFN}}. Interestingly, for both the \textit{Half-Cycle} and the \textit{Cycle} models, the use of the adversarial loss (red lines) reduces the amount of small errors ($\tau <1.00$). The amount of large errors ($\tau>1.00$) is similar to what is obtained without adversarial loss (green lines). We also observe that the performance gain of the \textit{Cycle} approach (solid lines) is spread uniformly over the whole range of errors. Similarly, adding the SSIM loss reduces uniformly the errors and leads to the best performing model.
  Concerning the Cityscapes dataset, the accuracy plot confirms the benefit of using our \textit{Cycle} approach. Similarly to the KITTI dataset, for both the \textit{Half-Cycle} and the \textit{Cycle} models, the use of the adversarial loss (red lines) reduces the amount of small errors ($\tau <1.00$) but the amount of large errors ($\tau>1.00$) is similar to what is obtained without adversarial loss (green lines). Nevertheless, the boost of the \textit{Cycle} approach is smaller than that on the KITTI dataset. When warping only the disparities, the \textit{Half-Cycle} and \textit{Cycle} models perform similarly but the use of a cycle improves the predictions when warping the feature maps (green and red lines). As observed on the KITTI dataset, the SSIM loss further reduces the prediction errors.
  
In addition to KITTI and Cityscapes, we report an ablation study on the ApolloScape dataset in Table~\ref{tab:ablation_apollo}. This dataset provides dense depth annotation  and therefore allows more accurate evaluation. 
We observe that, for both the \textit{Half-Cycle} and the \textit{Cycle}, our proposed feature alignment and sharing among stereo branches improves the perfomance. Moreover, it is clear that adversarial learning contributes to improving the results. In both \textit{Half-Cycle} and \textit{Cycle} settings, every evaluation metric shows an improvement. These observations are well in-line with the numbers reported on the KITTI and Cityscapes and further demonstrate the effectiveness of our approach.

\begin{figure*}[!t]
	\centering
	\includegraphics[width=.75\textwidth]{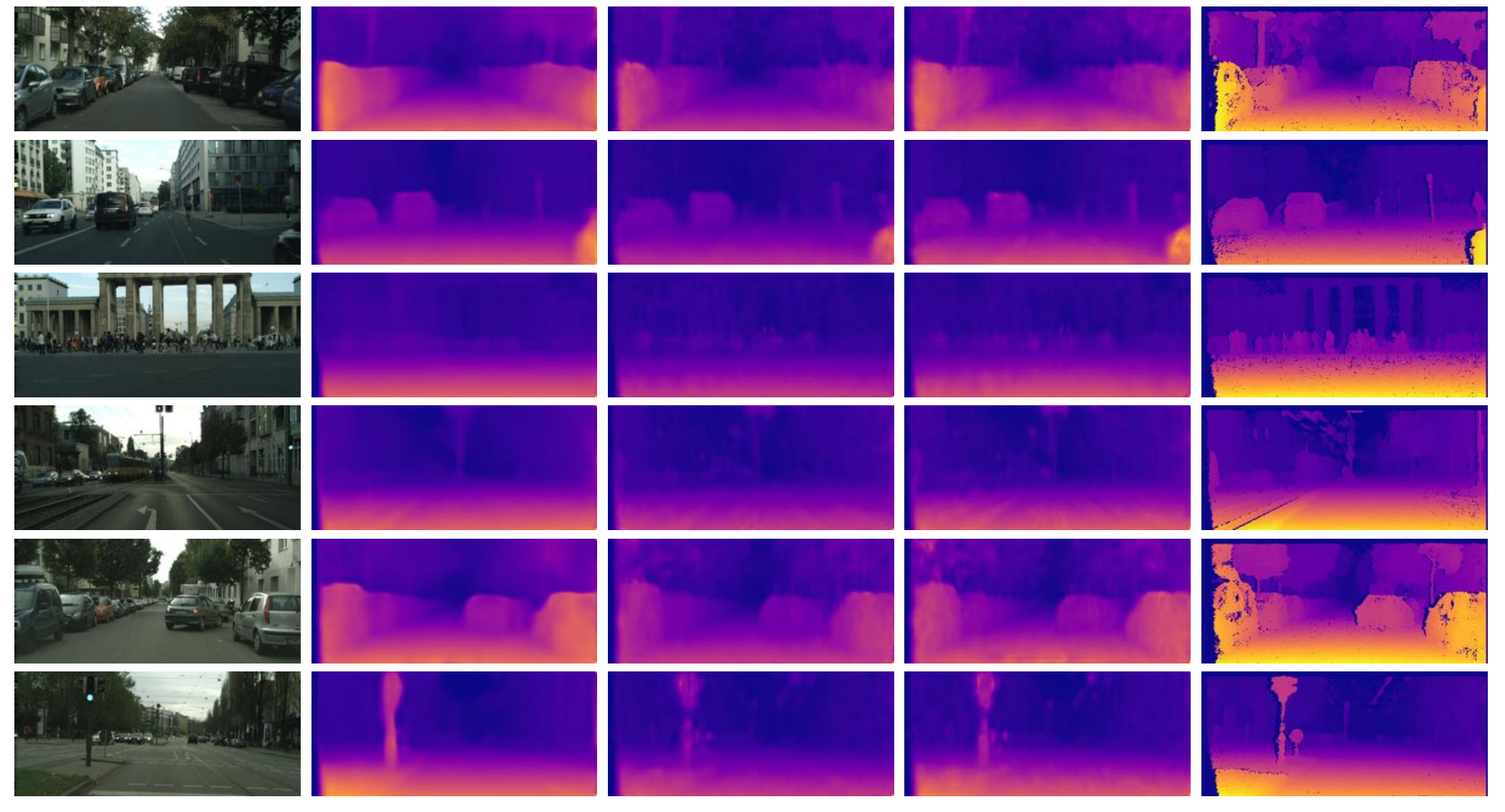}
	\put(-360,210){\scriptsize RGB Image}
	\put(-295,210){\scriptsize \textit{Half-Cycle}, $d$}
	\put(-230,210){\scriptsize \textit{Half-Cycle}, $d\oplus\xi$}
	\put(-160,210){\scriptsize \textit{Half-Cycle} + $\mathcal{L}_{GAN}$, $d\oplus\xi$}
	\put(-60,210){\scriptsize GT Depth Map}
	\vspace{-5pt}
	\caption{Qualitative comparison of different \textit{Stereo Half-Cycle} models on the Cityscapes testing dataset. The second column presents progressive disparity fusion, they are in general smoother but don't present the level of detail that we can find in third and fourth column where we have progressive feature fusion. Columns three and four present results from models learned with adversarial loss.}
	\label{fig:cityscapes}
	\vspace{-5pt}
\end{figure*}

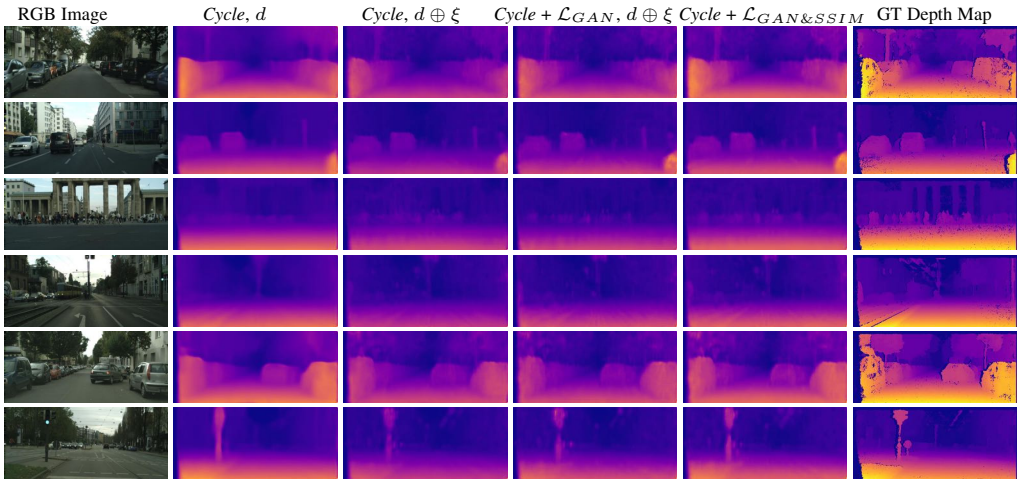
\begin{figure*}[!t]
	\centering
	\includegraphics[width=.75\textwidth]{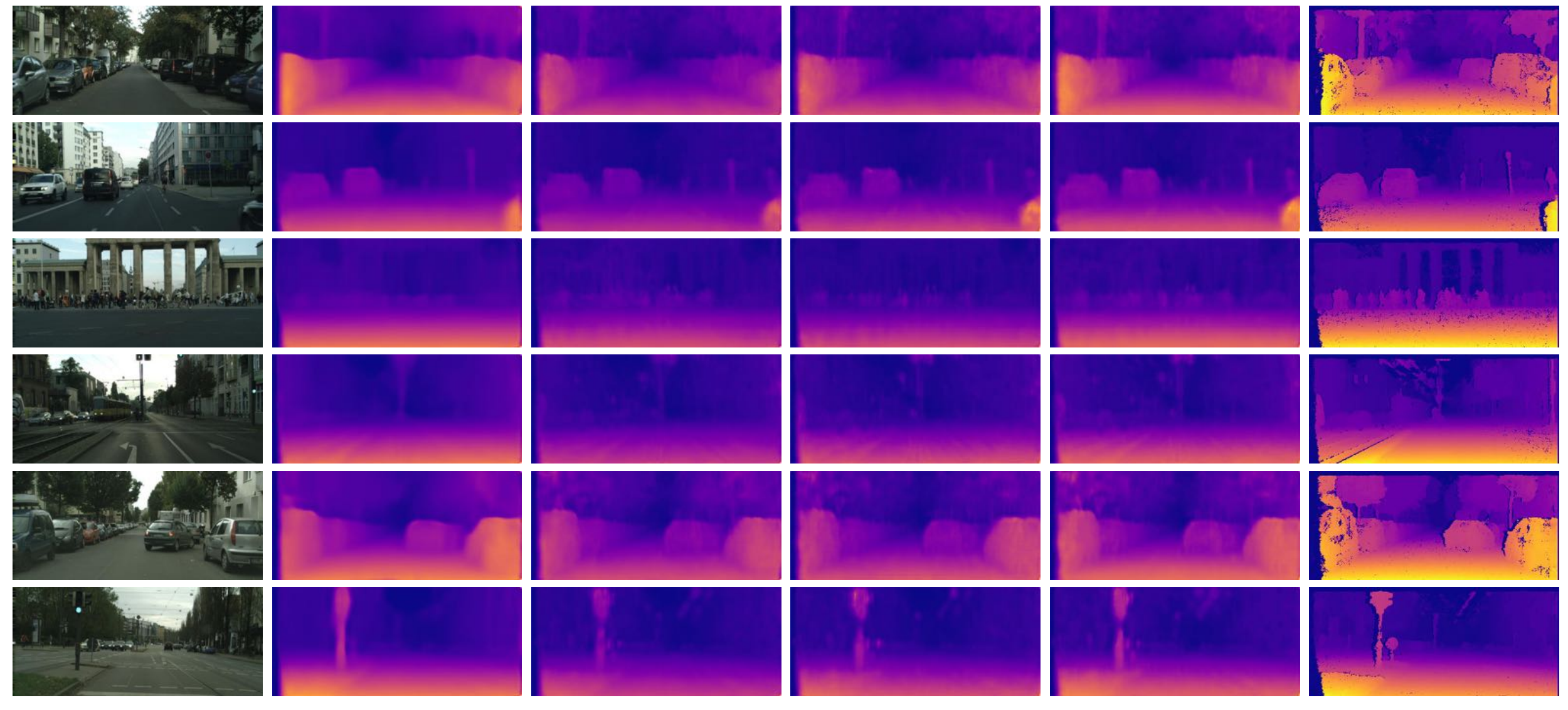}
	\put(-380,177){\scriptsize RGB Image}
	\put(-310,177){\scriptsize \textit{Cycle}, $d$}
	\put(-250,177){\scriptsize \textit{Cycle}, $d\oplus\xi$}
	\put(-200,177){\scriptsize \textit{Cycle} + $\mathcal{L}_{GAN}$, $d\oplus\xi$}
	\put(-130,177){\scriptsize \textit{Cycle} + $\mathcal{L}_{GAN \&SSIM}$} 
	\put(-55,177){\scriptsize GT Depth Map}
	\vspace{-5pt}
	\caption{Qualitative comparison of different \textit{Stereo Cycle} models on the Cityscapes testing dataset. Similarly to Fig.~\ref{fig:cityscapes} 
	in column two we present the results with progressive disparity fusion, while columns three, four and five have progressive features fusion.}
	\label{fig:cityscapes1}
\vspace{-5pt}
\end{figure*}

We also perform a qualitative comparison of the different baseline models with the proposed model. This qualitative evaluation is performed on the KITTI dataset and the results are shown in two figures in which we compare the \textit{Half-Cycle} models (Fig.~\ref{fig:kitti_half}) and the \textit{Cycle} models (Fig.~\ref{fig:kitti_cycle}) respectively. First, we observe that the \textit{Cycle} setting generates smoother disparity maps than the \textit{Half-Cycle} setting. In addition, when only the disparities are exchanged between the two streams of the\comRev{~\emph{PFN}}, we obtain very smooth predictions but with a low level of detail. When both the disparity and the feature maps are warped ($d \oplus \xi$ models), the predicted depth maps are more detailed and have sharper edges.
 In addition, by looking at the rows 6 and 8 of Figs. \ref{fig:kitti_half}) and \ref{fig:kitti_cycle}), we notice that the models without feature warping have difficulty in estimating the depth of nearby objects. Predicting accurate depth maps for these examples is challenging since the network needs to handle larger misalignment between the two input images. These examples illustrate the benefit of our proposed model which is better handle these difficult cases. Finally, by looking at the rows 1, 2, 4, 5 and 7, we can see that the models without adversarial loss underestimate the depth of the roads in foreground. Estimating the depth of the road is challenging since the image is almost uniform in these regions and the network cannot exploit the edges to estimate the disparity values. The fact that the GAN loss seems to help predicting the depth better for the uniform image regions may explain the reduction of small errors observed with the adversarial loss in Fig.\ref{fig:plots}.
Concerning the Cityscapes dataset, we observe a similar trend to the KITTI dataset by looking at the qualitative results reported in Figs. \ref{fig:cityscapes} and \ref{fig:cityscapes1}. The proposed\comRev{~\emph{PFN}} produces very smooth disparities but without much details when exchanging only the disparity maps between the two streams. For example the road sign in row 6 (Fig. \ref{fig:cityscapes} and \ref{fig:cityscapes1}) is barely distinguishable and the parked cars in row 5 appear in the disparity maps as a single continuous object. Models trained with feature and disparity warping ($d\oplus\xi$) improve the estimations allowing to capture better the details of the objects and the background for images in row 1, 2 and 5 (Fig. \ref{fig:cityscapes} and \ref{fig:cityscapes1}). 
This is further improved by the cycle setting and the adversarial training that, as shown in Fig. \ref{fig:cityscapes1}, produces more detailed disparities especially in challenging image areas such as those corresponding to the background.

\begin{table*}[!t]
\begin{center}
\resizebox{.8\textwidth}{!}{
		\begin{tabular}{ | >{}l | >{}c | >{}c || >{}c | >{}c | >{}c | >{}c | >{}c | >{}c | >{}c |}
			\toprule
			\multirow{2}{*}{Method} &\multirow{2}{*}{Discriminator} & \multirow{2}{*}{Warping} & Abs Rel & Sq Rel & RMSE & RMSE log & $\delta<1.25$ & $\delta<1.25^2$ & $\delta<1.25^3$ \\
			\cline{4-7} \cline{8-10}
    &&& \multicolumn{4}{>{}c|}{lower is better}& \multicolumn{3}{>{}c|}{higher is better} \\
\midrule
\textit{Half-Cycle} + ${L}_{gan}$ &$D$ shared &$d\oplus\xi$ &0.155 &1.398 &5.951 &\bf 0.245 &0.785 &0.919 &\bf 0.966\\
\textit{Half-Cycle} + ${L}_{gan}$ &$D_r$ and $D_l$ non-shared &$d\oplus\xi$ &\bf 0.148 &\bf 1.209 &\bf 5.827 &0.246 &\bf 0.789 &\bf 0.921 &\bf 0.966 \\
\bottomrule
\end{tabular}
}
\end{center}
\vspace{-10pt}
\caption{Ablation study: discriminator usage. We evaluate on KITTI the impact of sharing weights among discriminators.}
\label{tab:ablation_share-discr_kitti}
\vspace{-5pt}
\end{table*}

\begin{table*}[!t]
	\begin{center}
	\resizebox{.8\textwidth}{!}{
		\begin{tabular}{ | >{}l | >{}c | >{}c || >{}c | >{}c | >{}c | >{}c | >{}c | >{}c | >{}c |}
			\toprule
			\multirow{2}{*}{Method} &\multirow{2}{*}{Discriminator} & \multirow{2}{*}{Warping} & Abs Rel & Sq Rel & RMSE & RMSE log & $\delta<1.25$ & $\delta<1.25^2$ & $\delta<1.25^3$ \\
			\cline{4-7} \cline{8-10}
			&&& \multicolumn{4}{>{}c|}{lower is better}& \multicolumn{3}{>{}c|}{higher is better} \\
			\midrule
			\textit{Cycle} & w/o                        &$d\oplus\xi$ &0.149 &1.338 &5.837 &0.247 &0.792 &0.921 &0.966\\

			\textit{Cycle} + ${L}_{gan}$& $1^{st}$ half-cycle &$d\oplus\xi$ &0.144 &1.399 &5.794 &0.237 &0.801 &0.927 &0.969 \\
			\textit{Cycle} + ${L}_{gan}$& $2^{nd}$ half-cycle &$d\oplus\xi$ &0.141 &1.229 &\bf 5.685 &0.235 &0.804 &0.927 &0.969\\
			\textit{Cycle} + ${L}_{gan}$& Both half-cycles&$d\oplus\xi$ &\bf 0.137 &\bf 1.199 &5.721 &\bf 0.234 &\bf 0.806 &\bf 0.928 &\bf 0.970\\
			\bottomrule
		\end{tabular}
		}
	\end{center}
	\vspace{-10pt}
	\caption{Ablation study: discriminator usage. We evaluate on KITTI the impact of discriminators on different part of the architecture.}
	\label{tab:ablation_discr_fisrtorsecondhalf_kitti}
\vspace{-5pt}
\end{table*}

\noindent \textbf{Ablation study: exploiting resynthesized images.}
\begin{table*}[h]
\begin{center}
\resizebox{.8\textwidth}{!}{
 \begin{tabular}{|>{}l|>{}c|>{}c||>{}c|>{}c|>{}c|>{}c||>{}c|>{}c|>{}c|>{}c|}
\toprule
   \multirow{2}{*}{Method} & \multirow{2}{*}{Alignment}& \multirow{2}{*}{Warping} & \multicolumn{4}{>{}c||}{KITTI}& \multicolumn{4}{>{}c|}{ApolloScape} \\
   \cline{4-7} \cline{8-11}
    &&& Abs Rel & Sq Rel & RMSE & RMSE log & Abs Rel & Sq Rel & RMSE & RMSE log \\
\midrule
\textit{Half-Cycle}&w/o               &$d         $ &0.169 &1.503 &6.204 &0.265&0.501 &14.352 &16.134 &0.462  \\                
\textit{Half-Cycle}&w/o               &$d\oplus\xi$ &0.159 &1.468 &6.087 &0.251&0.491 &14.383 &16.366 &0.458   \\               
\textit{Half-Cycle} + ${L}_{gan}$&w/o &$d\oplus\xi$ & 0.151 &1.501 &5.905 &0.249& 0.468 & 13.107 & 15.023 &  0.445   \\                
\midrule                                                                                                                      
\textit{Half-Cycle}&With               &$d         $ &0.160 &1.374 &6.105 &0.261&0.485 &14.585 &16.098 &0.452  \\                
\textit{Half-Cycle}&With               &$d\oplus\xi$ &0.154 &1.260 &5.960 &0.254&0.469 &13.443 &15.779 &0.448  \\                
\textit{Half-Cycle} + ${L}_{gan}$&With &$d\oplus\xi$ & \bf 0.148 &\bf 1.209 &\bf 5.827 &\bf 0.246&\bf 0.446 &\bf 12.283 &\bf 14.600 &\bf 0.432  \\  
\bottomrule
\end{tabular}
}
\end{center}
\vspace{-10pt}
\caption{Ablation study: impact of feature alignment: we compare the results without alignment of disparities and features between the two stereo branches (upper half) and with alignment (bottom half). We conducted experiments on KITTI and Apolloscape. 
}
\label{tab:warping}
\vspace{-5pt}
\end{table*}
Our proposed model is designed in two \textit{Half-Cycle} blocks. The first reconstructs images that are used as input in the second \textit{Half-Cycle}.
In Table~\ref{tab:ablation_reg_kitti}, we compare two different approaches for exploiting the resynthetised images on both KITTI and Apolloscape.
We perform an experiment where we input in the second \textit{Half-Cycle} the right resynthetised image and the real left image.
This approach is compared to our approach where we input both synthesized images.
We observed that the performances decrease when we use $(\hat{I}_r,I_l)$ as input compared to our proposed \textit{Cycle} model that takes in input $(\hat{I}_r,\hat{I}_l)$. The performances are similar to our half-cycle baseline according to several metrics. It validates experimentally our design choice for the cycle inputs. \\
\textbf{Ablation study: discriminator usage.} In our model as described in Sec.\ref{sec:method}, we employ two discriminators, the first $D_r$ is applied to the right reconstructed images and the second $D_l$ to the left reconstructed images. We perform an experiment where we apply a single discriminator $D$ on both images. In Table~\ref{tab:ablation_share-discr_kitti}, we observed that using two discriminators ($D_r$ for $I_r$ and $D_l$ for $I_l$) is more effective.
\addnote[discr-discussion]{1}{A possible explanation is that the reconstruction errors are different between the two synthesized images. Indeed, when generating the right image, the pixels located in the right side of the objects are not visible in the left image because of self-occlusion. Therefore errors are larger on the right of objects. Symmetrically, when synthesizing the left image, errors are larger on the left side of objects. The difference distribution of reconstruction errors can explain why separated discriminators work better. } The second set of experiments concerning the use of disciminator focuses on the question of where the discriminators should be employed. Table~\ref{tab:ablation_discr_fisrtorsecondhalf_kitti} presents results obtained by using the discriminators at different locations on the KITTI dataset. These experiments illustrate the contribution of both the discriminators, on the first and the second \textit{Half-Cycles}. \\
\noindent \textbf{Ablation study: impact of feature alignment.}
In each stereo fusion layer, we align the feature maps as formulated in Eqn.~\eqref{eq:warp} in order to avoid misalignment issues . We now present experiments to measure the impact of this design choice. We evaluate a variant of our model without using our proposed disparity and feature alignment. More precisely, instead of aligning the feature maps before concatenation, we simply concatenate the disparities and the feature maps in each \emph{PFN} layer as in the the U-Net architecture. Results are reported in Table~\ref{tab:warping} for both KITTI and ApolloScape. These experiments demonstrate the impact of feature alignment and illustrate that a U-Net-based architecture performs better when handling feature misalignment.

\begin{table*}[!t]
	\begin{center}
	\resizebox{.75\textwidth}{!}{
		\begin{tabular}{ | l | c | c | c || c | c | c | c | c | c | c |}
			\toprule
			\multirow{2}{*}{Method} & \multirow{2}{*}{Sup}&\multirow{2}{*}{Camera}&\multirow{2}{*}{Video} & Abs Rel & Sq Rel & RMSE & RMSE log & $\delta<1.25$ & $\delta<1.25^2$ & $\delta<1.25^3$ \\
			\cline{5-8} \cline{9-11}
			&&&& \multicolumn{4}{c|}{lower is better}& \multicolumn{3}{c|}{higher is better} \\
			\midrule
			Saxena~\etal~\cite{saxena2006learning}&Y & M &N &0.280&-&8.734&-&0.601&0.820&0.926\\
			Eigen~\etal~\cite{eigen2014depth}&Y& M &N &0.190&1.515&7.156&0.270&0.692&0.899&0.967\\
			Liu~\etal~\cite{liu2016learning}&Y& M &N &0.202&1.614&6.523&0.275&0.678&0.895&0.965\\
			AdaDepth~\cite{adadepth}, 50m&Y& M &N &0.162&1.041&4.344&0.225&0.784&0.930&0.974\\
			Kuznietzov~\etal~\cite{kuznietsov2017semi}&Y& M&N&-&-&4.815&0.194&0.845& 0.957& \textit{0.987}\\
			Xu~\etal~\cite{xu2018monocular}  & Y & M &N & 0.132 & 0.911 & - & \textit{0.162} &0.804 &0.945 & 0.981 \\
            Jiang~\etal \cite{Jiang2018eccv} &Y& M &N  &0.131 &0.937 &5.032 &0.203 &0.827 &0.946 &0.981 \\
            Gan~\etal~\cite{Gan2018ECCV} &Y& M &N  &0.098 &0.666 &\textit{3.933} &0.173 &\textit{0.890} &0.964 &0.985 \\
            Guo~\etal~\cite{Guo2018ECCV} &Y& M &N &\textit{0.097} &\textit{0.653} &4.170 &0.170 &0.889 &\textit{0.967} &0.986 \\
						\midrule
						\midrule
			DF-Net \cite{DFNet2018eccv} &N& M& Y &0.150 &1.124 &5.507 &0.223 &0.806 &0.933 &0.973 \\
			Godard~\etal \cite{godard2018digging} &N & M& Y &\textit{0.115} &\textit{1.010} &\textit{5.164} &\textit{0.212} &\textit{0.858} &\textit{0.946} &\textit{0.974} \\
						\midrule
									\midrule    

			Zhou~\etal~\cite{zhou2017unsupervised}&N & M&N &0.208&1.768&6.856&0.283&0.678&0.885&0.957\\
			Garg~\etal~\cite{garg2016unsupervised}&N & M&N &0.169&1.08&5.104&0.273&0.740&0.904&0.962\\
		    Godard~\etal~\cite{godard2017unsupervised}&N & M&N &0.148&1.344&5.927& 0.247 &0.803&0.922 & 0.964\\
                        \midrule
                        Godard Stereo~\cite{godard2017unsupervised} &N & S&N &0.109 &1.120 &5.013 &0.205 &\bf 0.908 &\bf 0.954 &0.973\\
                        Pilzer~\etal~\cite{pilzer2018unsupervised} & N  & S&N &0.152 &1.388 &6.016 &0.247 &0.789 &0.918 &0.965\\
			DispNet~\cite{mayer2016large}&N & S&N &0.126 &0.919 &4.733 &0.200 &0.885 &\bf 0.954 &\bf 0.978\\       
    		MADNet~\cite{tonioni2019cvpr}&N & S&N &0.118 &1.090 &4.926 &0.213 &0.896 &\bf 0.954 &0.973\\

            PFN (Ours) &N & S&N & \bf 0.102 &\bf 0.802 &\bf 4.657 &\bf 0.196 & 0.882 &0.953 & 0.977\\
			\midrule
        	AdaDepth~\cite{adadepth}, 50m&N & M&N &0.203&1.734&6.251&0.284&0.687&0.899&0.958\\
			Pilzer~\etal~\cite{pilzer2018unsupervised}, 50m& N  & S&N &0.144 &1.007 &4.660 &0.240 &0.793 &0.923 &0.968\\
		
			DispNet~\cite{mayer2016large}, 50m&N & S&N &0.131 &0.712 &3.288 &0.189 &0.901 &0.961 &\bf 0.982\\
			MADNet~\cite{tonioni2019cvpr}, 50m&N & S&N &0.112 &0.753 &3.648 &0.200 &\bf 0.907 &\bf 0.958 &0.976\\       
			PFN (Ours), 50m&N & S&N &\bf 0.097 &\bf 0.586 &\bf 3.502 &\bf 0.185 &0.893 &0.957 & 0.979\\

			\bottomrule
		\end{tabular}
		}
	\end{center}
	\vspace{-10pt}
	\caption{Comparison with the state of the art. Training and testing are performed on the KITTI \cite{kitti} dataset. Supervised and semi-supervised methods are marked with Y in the supervision (Sup.) column, unsupervised methods with N. Monocular methods are marked M and binocular methods using stereo images at inference time are marked with S in the \emph{Camera}. Methods using a frame sequence in input and, thus, exploiting temporal information, are marked with Y in the \emph{Video} column. Numbers are obtained on Eigen test split with Garg image cropping. Depth predictions are capped at the common threshold of 80 meters, if capped at 50 meters we specify it. Best scores among static unsupervised methods are in bold. Best scores among other method categories are in italic.}
\label{tab:state-art_kitti}
\vspace{-2pt}
\end{table*}

\begin{table*}[!t]
\begin{center}
\resizebox{.75\textwidth}{!}{
 \begin{tabular}{|>{}l|>{}c|>{}c|>{}c||>{}c|>{}c|>{}c|>{}c|>{}c|>{}c|>{}c|}
\toprule
   	\multirow{2}{*}{Method} & \multirow{2}{*}{Sup}&\multirow{2}{*}{Camera}&\multirow{2}{*}{Video} & Abs Rel & Sq Rel & RMSE & RMSE log & $\delta<1.25$ & $\delta<1.25^2$ & $\delta<1.25^3$ \\
			\cline{5-8} \cline{9-11}
			&&&& \multicolumn{4}{>{}c|}{lower is better}& \multicolumn{3}{>{}c|}{higher is better} \\
			\midrule
Eigen~\cite{eigen2014depth}&N & M & N &1.006&16.840 &20.620&1.156&0.229&0.435&0.583\\
Godard~\cite{godard2017unsupervised}&N & M &N &0.432 &12.199 &14.497 &0.426 &0.591 &0.832 &0.911\\
			\midrule
Godard Stereo~\cite{godard2017unsupervised}&N & S&N &0.397 &10.468 &13.865 &0.402 &0.594 &\bf 0.848 &\bf 0.933\\
Pilzer~\etal~\cite{pilzer2018unsupervised}&N & S&N  &0.473 &13.660 &15.556 &0.451 &0.558 &0.800 &0.893\\
PFN (Ours)&N & S&N  &\bf 0.387 &\bf 10.097 &\bf 13.449 &\bf 0.396 &\bf 0.669 &0.843 &0.915\\ 
\bottomrule
\end{tabular}
}
\end{center}
\vspace{-10pt}
\caption{Comparison with the state of the art on ApolloScape. We compare out model trained on ApolloScape with those of Godard~\etal~\cite{godard2017unsupervised} and Pilzer~\etal~\cite{pilzer2018unsupervised} using the code provided by the authors.}
\label{tab:sota_apollo}
\vspace{-8pt}
\end{table*}

\begin{figure*}[!t]
\centering
\includegraphics[width=.85\textwidth]{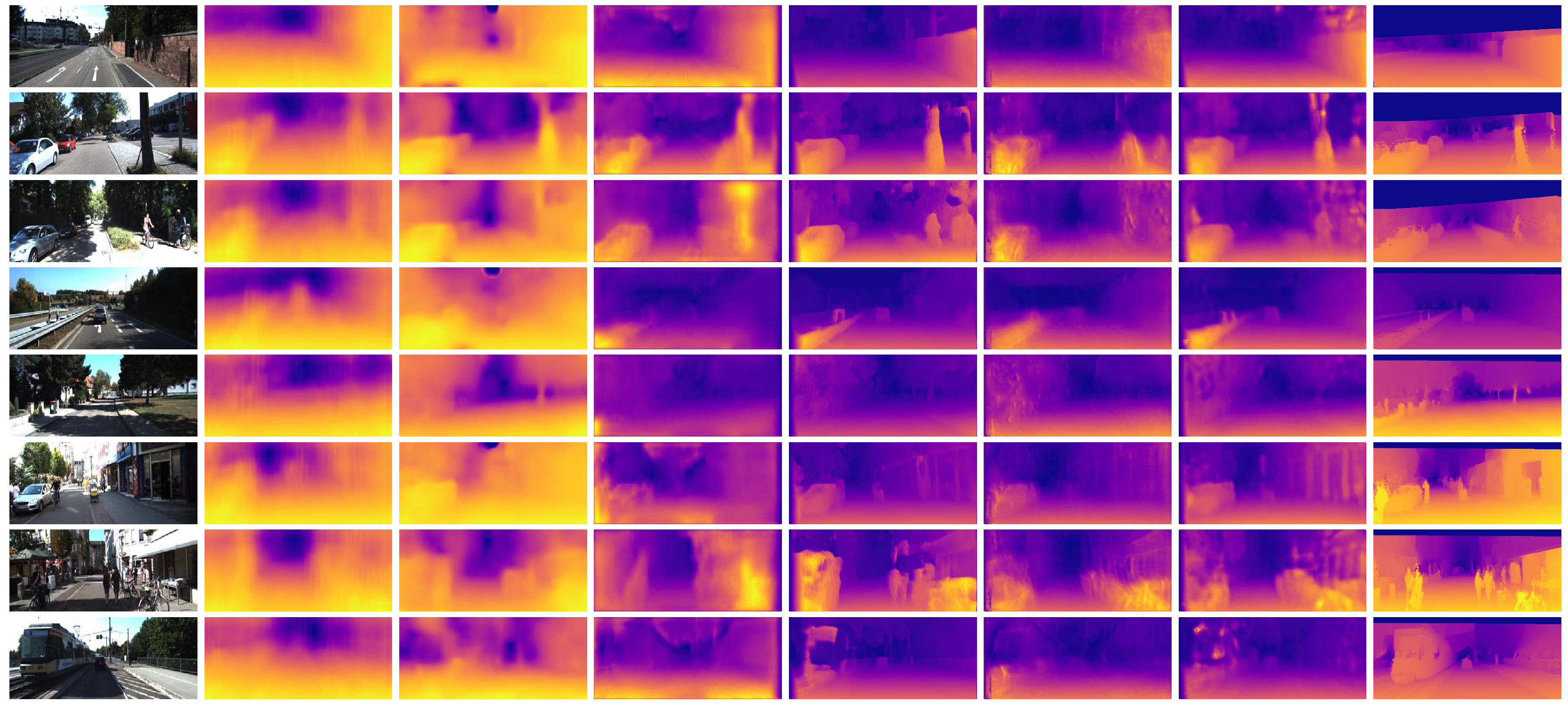}
	\put(-425,200){\scriptsize RGB Image}
	\put(-375,200){\scriptsize Eigen~\etal~\cite{eigen2014depth}}
	\put(-320,200){\scriptsize Zhou~\etal~\cite{zhou2017unsupervised}}
	\put(-270,200){\scriptsize Garg~\etal~\cite{garg2016unsupervised}}
	\put(-215,200){\scriptsize Godard~\etal~\cite{godard2017unsupervised}}
	\put(-160,200){\scriptsize Pilzer~\etal~\cite{pilzer2018unsupervised}}
	\put(-90,200){\scriptsize Ours }
	\put(-50,200){\scriptsize GT Depth Map}
\vspace{-4pt}
 \caption{Qualitative comparison with different competitive approaches with both supervised and unsupervised settings on the Eigen test set of KITTI dataset. The sparse groundtruth depth maps are filled with bilinear interpolation for better visualization.}
 \label{fig:qualComp}
 \vspace{-10pt}
\end{figure*}
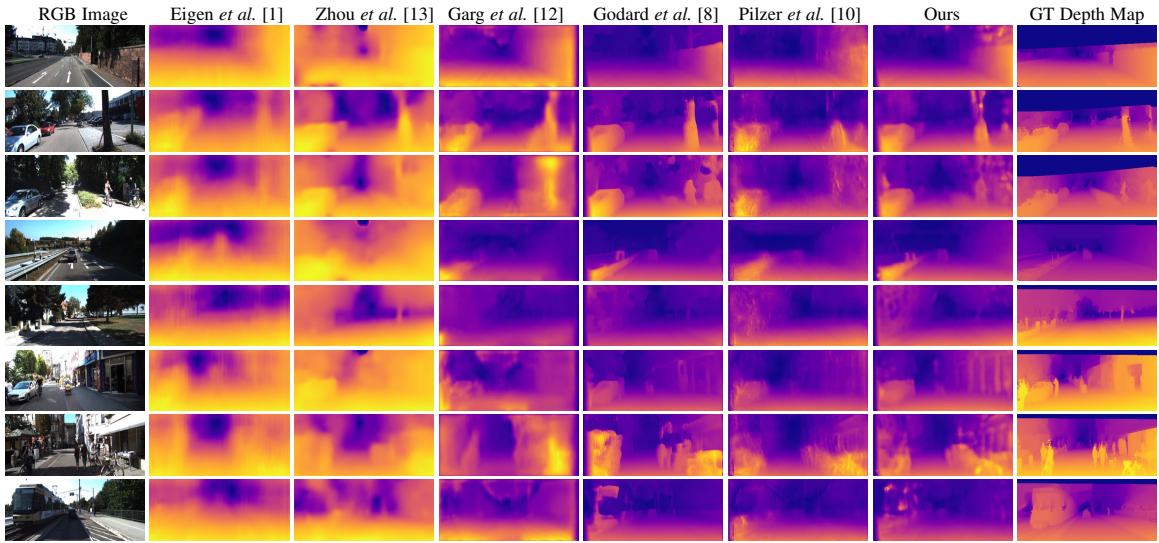

\subsection{Comparison with the State of the Art}
In Table~\ref{tab:state-art_kitti}, we compare the proposed full model with several state-of-the-art methods, including the ones with the supervised setting, \ie~Saxena~\etal~\cite{saxena2006learning}, Eigen~\etal~\cite{eigen2014depth}, Liu~\etal~\cite{liu2016learning}, AdaDepth~\cite{adadepth}, Kuznietzov~\etal~\cite{kuznietsov2017semi}, Xu~\etal~\cite{xu2018monocular}, Jiang~\etal \cite{Jiang2018eccv}, Gan~\etal~\cite{Gan2018ECCV} and Guo~\etal~\cite{Guo2018ECCV}, and the ones with the unsupervised setting, \ie~Zhou~\etal~\cite{zhou2017unsupervised}, AdaDepth~\cite{adadepth}, Garg~\etal~\cite{garg2016unsupervised},			DispNet~\cite{mayer2016large}, MADNet~\cite{tonioni2019cvpr} and Godard~\etal~\cite{godard2017unsupervised}. As far as we know, there are not quantitative results presented in the existing works on the Cityscapes dataset and for this reason, we perform the comparison on the KITTI dataset. 
These results further demonstrate the potential of unsupervised training for depth estimation. Note that we do not include the recent work in~\cite{Yang2018ECCV} in Table~\ref{tab:state-art_kitti} as a different experimental setup is considered (different training/test split). Furthermore in~\cite{Yang2018ECCV} additional information (ego-motion information) is exploited for depth prediction.

For comparison with the unsupervised methods, we outperform previous methods, according to four metrics: Abs Rel, Sq Rel, RMSE and RMSE log. In particular, we outperform the binocular methods DispNet~\cite{mayer2016large}, MADNet~\cite{tonioni2019cvpr} and the method proposed by Godard~\etal~\cite{godard2017unsupervised}. According to accuracy metrics, we are on par with theses recent approaches. These results illustrate the benefit of our approach. \addnote[madnet-dispnet]{1}{Note that we compared with DispNet~\cite{mayer2016large}, MADNet~\cite{tonioni2019cvpr} since they are two recent architectures for stereo matching with a code that is publicly available and ready-to-use. Even though, these architectures are trained with supervision in the original work, we report the performances obtained when training in the self-supervised setting.}~   
Concerning AdaDepth~\cite{adadepth}, we must mention that their approach is, to some extent, related to our approach since they employ adversarial learning in a context of domain adaptation with extra synthetic training data. Therefore, the better performance of our model illustrates the superiority of the proposed cycle-based data-augmentation approach compared to their use of synthetic data. When we consider the methods that use several frames at training and test time, only \cite{godard2018digging} performs better.
Regarding ApolloScape dataset, to the best of our knowledge, we are the first to benchmark depth estimation methods on this dataset. We compare our approach with the competitive unsupervised models proposed in~\cite{godard2017unsupervised,pilzer2018unsupervised} using publicly available codes. Results are presented in Table~\ref{tab:sota_apollo}, our proposed model improves by a large margin over~\cite{pilzer2018unsupervised} and the monocular model of~\cite{godard2017unsupervised}, and, importantly, it improves also with respect to the binocular model of~\cite{godard2017unsupervised} according to five metrics over seven.

  The conclusion drawn in this quantitative comparison are confirmed by the qualitative evaluation reported in Fig~\ref{fig:qualComp}. Compared to our previous work \cite{pilzer2018unsupervised}, we see that our model estimates better the edges of the objects that appear in the image. For instance in the second row, we can distinguish better the shape of the trunk of the tree in foreground. The same remark stands for the reconstruction of the cars in the $2^{nd}, 3^{rd}, 4^{th}$ and $6^{th}$ rows. Comparing with \cite{godard2017unsupervised}, we distinguish similarly the edges of the objects but we estimate better the depth of large horizontal regions of the images. For instance, the depth of the road is much better estimated by our proposed model. This is especially true for the rows 1, 5 and 6 in which\cite{godard2017unsupervised} underestimates the depth of the road. It can be explained by the difficulty of handling large displacements between the left and the right object when the image region does not contain many edges.


In addition to the several novelties presented in this work with respect to our previous work \cite{pilzer2018unsupervised}, the newly proposed model has fewer parameters and a lower training complexity. The best performing model of \cite{pilzer2018unsupervised} consisted of seven main blocks, an encoder extracting the features from the images, four decoders, trained to reconstruct disparities and two discriminators, one for the right stereo view and one for the left stereo view. This complex model is trained iteratively to guarantee a good starting point for fine-tuning the network. Despite the good performance, the model proposed in our previous work has a complex optimization process.

\section{Conclusions}
\label{sec:concl}
We have presented a novel approach for unsupervised deep learning for the depth estimation employing a cycle structure. This new approach uses cycle consistency such that the network does not only learn from the training set images but also from the images generated in the first half-cycle. In addition, we proposed a generative deep network model specifically designed for binocular stereo depth estimation. By combining a refinement approach with a multi-scale strategy we improve the quality of the predicted depth map. It is worth noticing that although tested in the unsupervised setting, the proposed \emph{PFN} can be also used in a supervised stereo scenario. In this work we decided to focus on the unsupervised setting because of the aforementioned practical advantages. However, monocular depth estimation methods can also benefit from the proposed adversarial approach.
Extensive experiments were conducted on three publicly available datasets \ie the popular KITTI, Cityscapes and ApolloScape datasets. Our results demonstrate the effectiveness of the proposed model, which is competitive with state of the art approaches for unsupervised depth estimation.
As future works we plan to evaluate the proposed \emph{PFN} to other prediction tasks in which the misalignment between the input images may affect the performance, \ie optical flow estimation or video frame interpolation. 



%


\ifCLASSOPTIONcompsoc
  \section*{Acknowledgments}
\else
  \section*{Acknowledgment}
\fi

The authors would like to thank the NVIDIA Corporation for the donation of the
GPUs used in this project. This work was carried out under the "Vision and Learning joint Laboratory" between FBK and UNITN.

\ifCLASSOPTIONcaptionsoff
  \newpage
\fi



\bibliographystyle{IEEEtran}

%

%
\vspace{-30pt}
\begin{IEEEbiography}[{\includegraphics[width=1in,height=1.25in,clip,keepaspectratio]{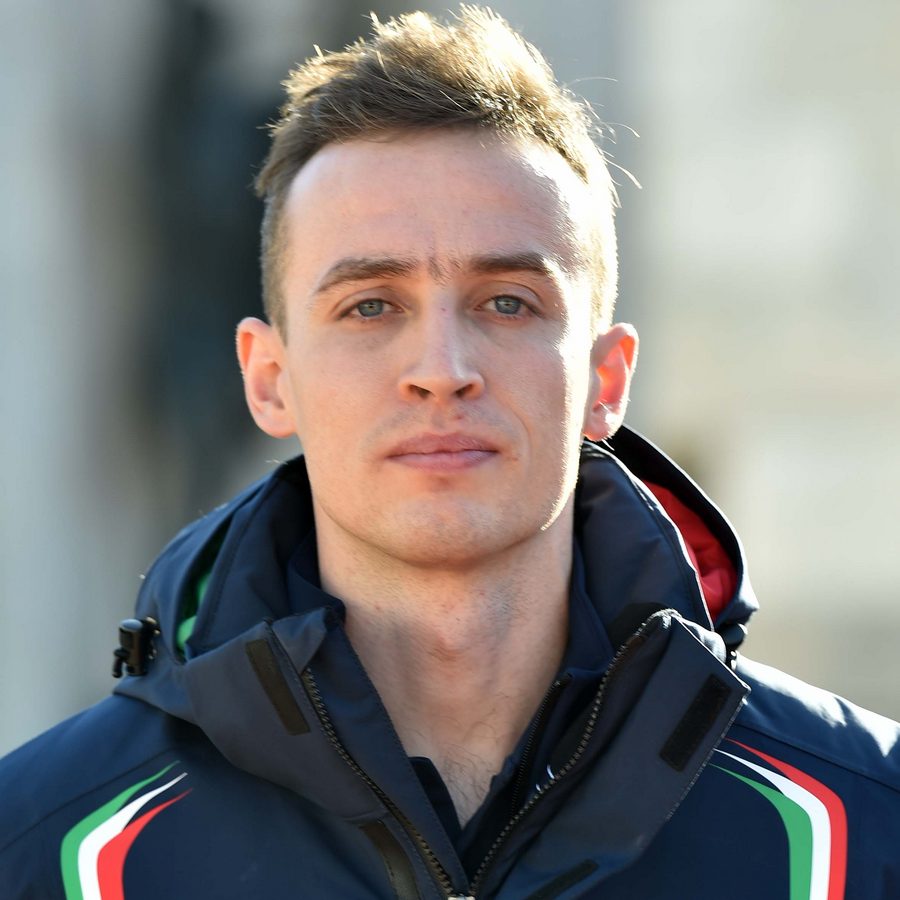}}]{Andrea Pilzer} received his M.Sc degree in Telecommunications Engineering from University of Trento in 2016. After he joined the MHUG research group at the Univeristy of Trento as PhD student under the supervision of prof. Nicu Sebe and prof. Elisa Ricci. His research interests are in deep learning for dense regression tasks and social network visual content understanding. 
\end{IEEEbiography}
\vspace{-35pt}
\begin{IEEEbiography}[{\includegraphics[width=1in,height=1.25in,clip,keepaspectratio]{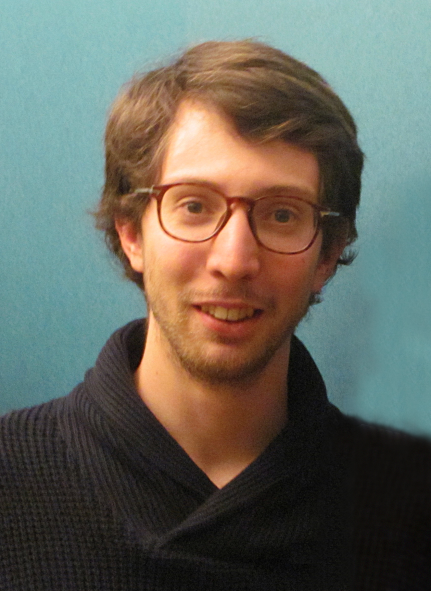}}]{St\'{e}phane Lathuili\`{e}re} is an assistant professor at T\'{e}l\'{e}com ParisTech,
France. He received the M.Sc. degree in applied mathematics and
computer science from Grenoble Institute of Technology in 2014. He worked
towards his Ph.D. in the
Perception Team at Inria under the supervision of Dr. Radu Horaud,
and obtained it from Université\'{e} Grenoble Alpes (France) in 2018. 
He was a Postdoc researcher at the  Univeristy of Trento. His
research interests cover deep models for regression, generation and domain adaptation.
\end{IEEEbiography}

\vspace{-35pt}
\begin{IEEEbiography}[{\includegraphics[width=1in,height=1.25in,clip,keepaspectratio]{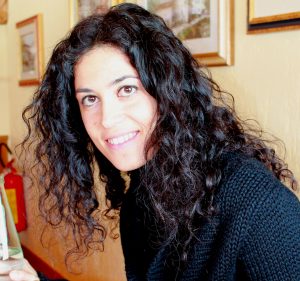}}]{Elisa Ricci}
received the PhD degree from the
University of Perugia in 2008. She is an assistant
professor at the University of Perugia and
a researcher at Fondazione Bruno Kessler. She
has since been a post-doctoral researcher at
Idiap, Martigny, and Fondazione Bruno Kessler,
Trento. She was also a visiting researcher at
the University of Bristol. Her research interests
are mainly in the areas of computer vision and
machine learning. She is a member of the IEEE.
\end{IEEEbiography}
\vspace{-35pt}
\begin{IEEEbiography}[{\includegraphics[width=1in,height=1.25in,clip,keepaspectratio]{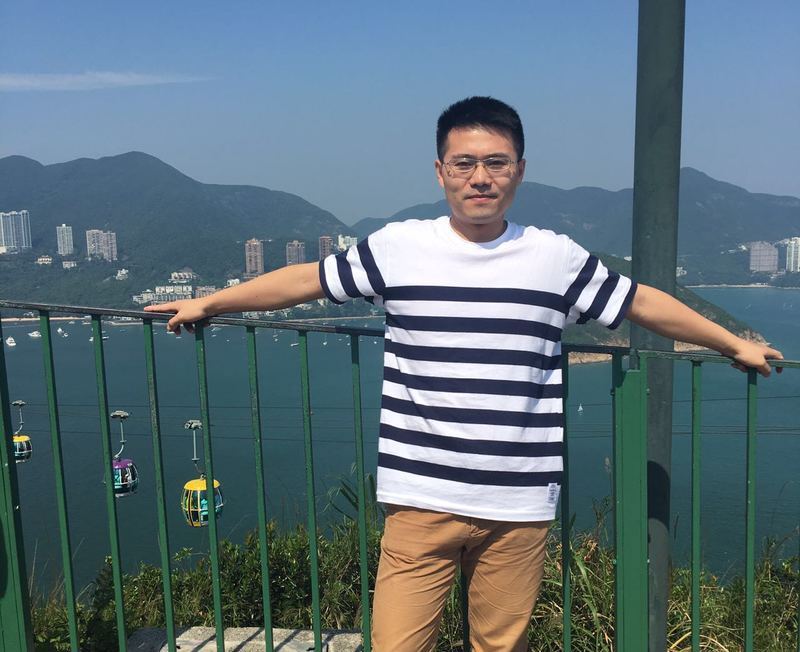}}]{Dan Xu} is a Postdoc researcher in Visual Geometric Group at the University of Oxford. He received the Ph.D. Computer Science from the University of Trento. 
His research
focuses on computer vision, multimedia and machine
learning. Specifically, he is interested in
deep learning, structured prediction and cross-modal
representation learning and the applications to 2D/3D scene understanding
tasks. He received the Intel best scientific paper award at ICPR
2016.

\end{IEEEbiography}
\vspace{-35pt}
\begin{IEEEbiography}[{\includegraphics[width=1in,height=1.25in,clip,keepaspectratio]{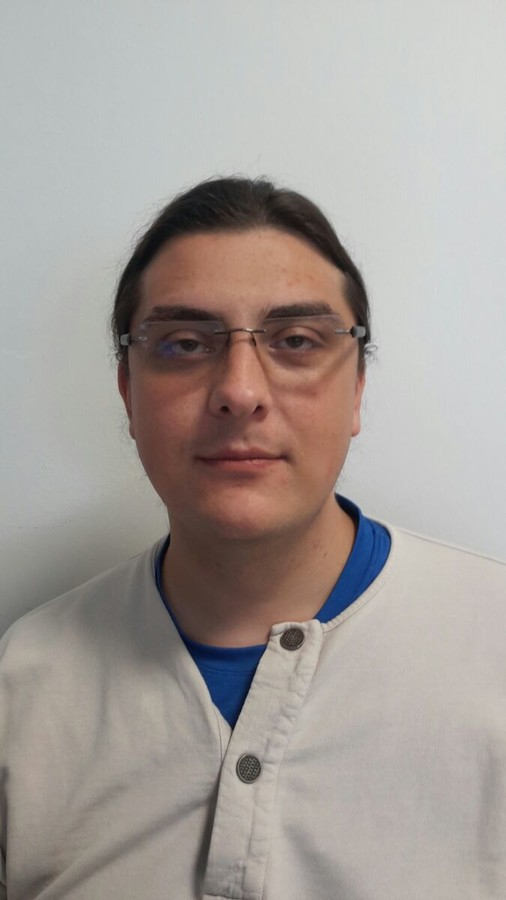}}]{Mihai Marian Puscas} is a Video Research Intern at the Huawei Ireland Research Center. His research focuses on learning in scarce data regimes, more specifically few-shot learning, life-long learning and unsupervised image and video understanding. He received his Ph.D. in Computer Science from the University of Trento.

\end{IEEEbiography}
\vspace{-30pt}
\begin{IEEEbiography}[{\includegraphics[width=1in,height=1.25in,clip,keepaspectratio]{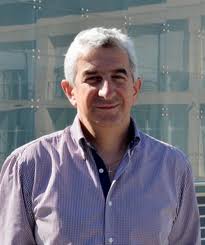}}]{Nicu Sebe}
is Professor with the University of
Trento, Italy, leading the research in the areas
of multimedia information retrieval and human
behavior understanding. He was the General
Co- Chair of the IEEE FG Conference 2008 and
ACM Multimedia 2013, and the Program Chair
of the International Conference on Image and
Video Retrieval in 2007 and 2010, ACM Multimedia
2007 and 2011. He was the Program Chair
of ICCV 2017 and ECCV 2016, and a General
Chair of ACM ICMR 2017. He is a fellow of the
International Association for Pattern Recognition.
\end{IEEEbiography}



\end{document}